\documentclass[10pt,twocolumn,letterpaper]{article}

\pdfoutput=1
\usepackage{cvpr}
\usepackage{times}
\usepackage{epsfig}
\usepackage{graphicx}
\usepackage{amsmath}
\usepackage{amssymb}
\usepackage{breqn}
\usepackage{array}
\usepackage{subfigure}
\usepackage{enumitem}
\usepackage{placeins}
\usepackage[normalem]{ulem}

\makeatletter
\newcommand\footnoteref[1]{\protected@xdef\@thefnmark{\ref{#1}}\@footnotemark}
\makeatother

\newcommand{\eat}[1]{}
\newcommand{\rev}[1]{{#1}} 
\renewcommand{\sout}[1]{}

\newcommand{\mymodel}{SemStyle\xspace}
\newcommand{\termgenr}{term generator\xspace}
\newcommand{\termgen}{{\it term generator}\xspace}

\newcommand{\langgenr}{language generator\xspace}
\newcommand{\langgen}{{\it language generator}\xspace}

\newcommand{\styleterm}{{\it target-style term}\xspace}

\newcommand{\imemb}{image embedder\xspace}
\newcommand{\sentemb}{{\it sentence embedder}\xspace}
\newcommand{\sentgen}{{\it sentence generator}\xspace}

\DeclareMathOperator*{\argmax}{argmax}

\usepackage[breaklinks=true,bookmarks=false]{hyperref}

\cvprfinalcopy 

\def\cvprPaperID{4057} 

\ifcvprfinal\fi

\usepackage{cancel} 
\definecolor{navy}{rgb}{0.1, 0.1, 0.8}
\definecolor{gray}{rgb}{0.4, 0.4, 0.4}
\definecolor{myblue}{rgb}{.8, .8, 1}
\definecolor{olive}{rgb}{0.1, 0.5, 0.1}
\definecolor{myyellow}{RGB}{255,215,0}

\newcommand{\titlemoveup}{\vspace{-1.0mm}}
\newcommand{\textmoveup}{\vspace{-0.5mm}}         	
\newcommand{\eqmoveup}{\vspace{-2.0mm}} 
\newcommand{\captionmoveup}{\eqmoveup\vspace{-1.8mm}}   

\begin{document}

\title{SemStyle: Learning to Generate Stylised Image Captions using Unaligned Text}

\author{Alexander Mathews$^{*\dagger}$, Lexing Xie$^{*\dagger}$, Xuming He$^\ddagger$\\
Australian National University$^*$, Data to Decision CRC$^\dagger$, ShanghaiTech University$^\ddagger$\\
{\tt\small alex.mathews@anu.edu.au, lexing.xie@anu.edu.au, hexm@shanghaitech.edu.cn}
}

\maketitle
\thispagestyle{empty}

\begin{abstract}\textmoveup
Linguistic style is an essential part of written communication, with the power to affect 
both clarity and attractiveness. With recent advances in vision and language,
we can start to tackle the problem of generating image captions that are both visually grounded and appropriately styled. Existing approaches either require styled training captions aligned to images or generate captions with low relevance.
We develop a model that learns to generate visually relevant styled captions from a large corpus of styled text without aligned images. The core idea of this model, called {\it\mymodel}, is to separate semantics and style.
One key component is a novel and concise semantic term representation 
generated using natural language processing techniques and frame semantics. 
In addition, we develop a unified language model that decodes sentences 
with diverse word choices and syntax for different styles.
Evaluations, both automatic and manual, show captions from {\it\mymodel} 
preserve image semantics, are descriptive, and are style shifted. 
More broadly, this work provides possibilities to learn richer image descriptions 
from the plethora of linguistic data available on the web. 
\end{abstract}

\titlemoveup\titlemoveup
\section{Introduction}
\label{sec:intro}
\textmoveup

An image can be described in different styles, for example, from a first-person or third-person perspective, with a positive or neutral sentiment, in a formal or informal voice. 
Style is an essential part of written communication that reflects personality~\cite{pennebaker1999linguistic},
influences purchasing decisions~\cite{ludwig2013more} and fosters social interactions~\cite{danescu2011mark,niederhoffer2002linguistic}.
The analysis of linguistic styles~\cite{danescu2011mark,pavlick2015inducing} and generating natural language descriptions~\cite{vinyals2015show} are two fast developing topics in language understanding and computer vision, but an open challenge remains at their intersection: writing a visually relevant sentence in a given style. \rev{Incorporating style into image captions will help to communicate image content clearly, attractively, and in a way that is emotionally appropriate.} For this work, we focus on writing one sentence to describe an image.

Style traditionally refers~\cite{pennebaker1999linguistic} to linguistic aspects other than the message content. It can be defined in terms of a fixed set of attributes~\cite{pavlick2015inducing, ficler2017controlling} such as formality and complexity, or implicitly with a document collection from a single author~\cite{Stamatatos2009} or genre~\cite{Kiros2015b}.
The early works on stylistic image captioning model word changes~\cite{mathews2016senticap}, and transformation of word embeddings~\cite{gan2017stylenet}, but do not explicitly separate content and style.
These works also require manually created, style specific, image caption datasets~\cite{mathews2016senticap,gan2017stylenet}, and are unable to use large collections of styled text that does not describe images. We aim to address three gaps in the current solutions. 
The first is human-like style transfer: using large amounts of unrelated text 
in a given style to compose styled image captions. 
\rev{This is in contrast to existing systems that require aligned images and styled text.} The second is representing an image so that the semantics are preserved while allowing flexible word and syntax use.
The third is ensuring stylistic text remains descriptive and relevant to the image.

\begin{figure}
\centering
\includegraphics[width=0.5\textwidth]{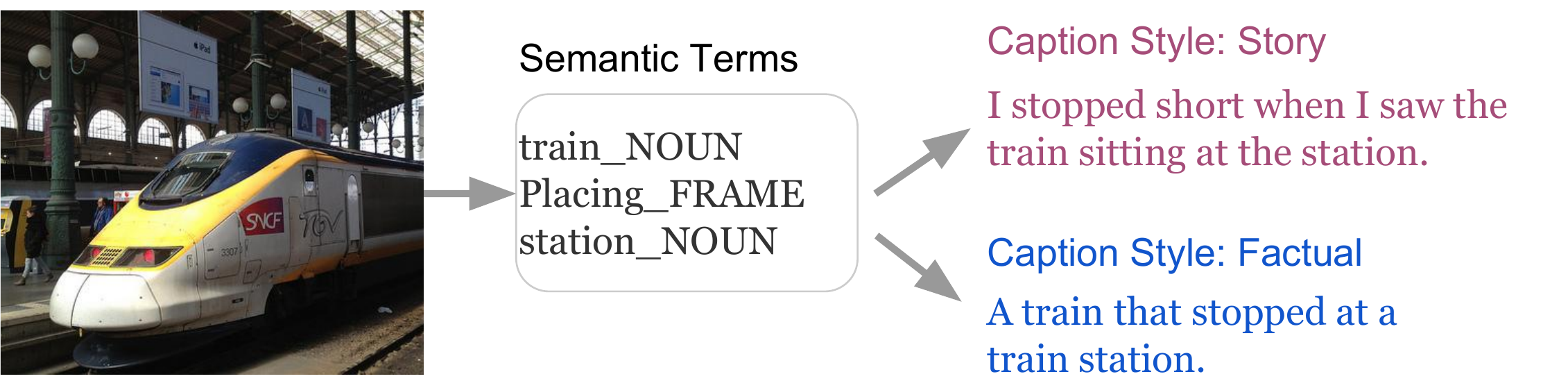}
\caption{{\it\mymodel} distils an image into a set of semantic terms, which are then used to form captions of different styles.}
\label{fig:intro}\captionmoveup
\end{figure}\textmoveup

We develop a model, dubbed {\it\mymodel}, for generating stylistically interesting and semantically relevant image captions by learning from a large corpus of stylised text without aligned images.
Central to our approach is a separation of concerns regarding semantic relevance and style. We propose a novel semantic terms representation that is concise and promotes flexibility in word choice. This term representation consists of normalised words with part-of-speech tag, and verbs generalised using the lexical database FrameNet~\cite{Baker1998}. Further, we develop a \termgen for obtaining a list of terms related to an image, and a \langgen that decodes the ordered set of semantic terms into a stylised sentence. The \termgen is trained on images and terms derived from factual captions. The \langgen is trained on sentence collections and is conditioned to generate the desired style. As illustrated in Figure~\ref{fig:intro}, the \termgen produces train\_NOUN, Placing\_FRAME, station\_NOUN from the image, and the \langgen produces sentences of different styles from this set of terms. Evaluated on both MSCOCO~\cite{chen2015microsoft} and a corpus of romance novels~\cite{Zhu2015}, the \mymodel system produced distinctively styled captions in 58.8\% of cases, while retaining visual semantics as judged by the SPICE metric~\cite{Anderson2016}. Evaluated subjectively by the crowd, \mymodel achieved an average descriptiveness of 2.97 (out of 4, larger is more descriptive), which is competitive with a purely descriptive baseline at 2.95. Since this descriptive baseline is the basis of the \termgen we can conclude that \mymodel retains the descriptive accuracy of the underlying semantic model. Moreover, 41.9\% of captions from \mymodel were judged to be telling a story about the associated image. The main contributions of this paper are as follows: 

\begin{itemize}[noitemsep,topsep=0pt,parsep=0pt,partopsep=0pt]
\item A concise semantic term representation for image and language semantics, implemented with a neural-network based \termgen.
\item A method that uses \rev{semantic terms} to generate relevant captions with wording flexibility.
\item A training strategy for learning to mimic sentence-level style using both styled and descriptive corpora.
\item Competitive results in human and automatic evaluations with existing, and two novel, automated metrics for style. Dataset, models and evaluation results are released online~\footnote{\label{ft:data}\url{https://github.com/computationalmedia/semstyle}}. 
\end{itemize}

\titlemoveup
\section{Related work}
\label{sec:related_work}
\textmoveup

This work is closely related to recent work on recognising linguistic style, captioning images with additional information, and the new topic of generating stylistic captions. 

The problem of identifying writing style to identify authors has received much interest.
Many features have been proposed~\cite{Stamatatos2009} to disentangle style and content
including: lexical features~\cite{Argamon2003,Argamon2005,VanHalteren2005}, bag of functional words~\cite{Argamon2005} and Internet slang terms~\cite{Argamon2003}. Paraphrasing, and word choice are shown to be important indicators of style~\cite{xu2012paraphrasing,pavlick2015inducing}. In online communities, writing style is shown to be indicative of personality~\cite{schwartz2013personality}, vary across different online fora, and change throughout a discussion~\cite{pavlick2016empirical}.
Synthesising text in a particular style is an emerging problem in natural language generation~\cite{ghazvininejad2016poetry, kiddon2016globally}, 
but the quality of results is limited by the size of parallel text collection in two different styles~\cite{jhamtani2017shakespear}. In other cases the semantic content is not controlled and so may not be relevant to the subject such as an image or movie~\cite{ficler2017controlling}. \sout{We address both problems in this work.} \rev{In this work we address both problems with respect to stylised image captioning.}

Current state-of-the-art image captioning models consist of a Convolutional Neural Network~\cite{LeCun1998} (CNN) for object detection, and a Recurrent Neural Network~\cite{Elman1990,hochreiter1997long} (RNN) for caption generation~\cite{donahue2015long,Karpathy2015,Kiros2014a,mao2014deep,vinyals2015show,liu2017improved, yao2017boosting, pedersoli2017areas, tan2016phi}. These two components can be composed and learnt jointly through back-propagation. Training requires hundreds of thousands of aligned image-caption pairs, 
and the resulting captions reflect the purely descriptive style of the training data. 
Generalising image caption generators beyond the standard caption datasets (such as MSCOCO~\cite{chen2015microsoft}) is thus of interest. Some captioning systems are designed to generalise to unseen objects~\cite{mao2015learning,anne2016deep,venugopalan2017captdiv}; 
Luong~\etal~\cite{Luong2016} exploit other linguistic tasks via multi-task sequence learning; Venugopalan~\cite{venugopalan:emnlp16} show that additional text improves the grammar of video captions. 
While this work leverages the general idea of multi-task learning, our specific proposal for separating semantics and style is new, as is our model for generating stylistic text. 

Neural-storyteller~\cite{Kiros2015b} generates styled captions by retrieving candidate factual captions~\cite{Kiros2014a}, embedding them in a vector space~\cite{Kiros2015a}, shifting the embeddings by the mean of the target style (e.g. romance novels) before finally decoding. The resulting captions are representative of the target style but only loosely related to the image.
Our earlier SentiCap system~\cite{mathews2016senticap} generates captions expressing positive or negative sentiment. Training employs a switching RNN to adapt a language decoder using a small number of training sentences. This approach needs an aligned dataset of image captions with sentiment, and word-level annotations to emphasize words carrying sentiment. The StyleNet system~\cite{gan2017stylenet} uses a factored weight matrix to project word embeddings. Style is encoded in this factored representation while all other parameters are shared across different styles. This approach uses styled image-captions pairs for learning.
To best of our knowledge, no stylistic image caption system exists that is grounded on images, adapts both word choice and syntactic elements, and is able to learn on large collections of text without paired images.


\section{Our Approach}
\label{sec:model}

\begin{figure*}
	\centering
	\includegraphics[width=0.9\textwidth]{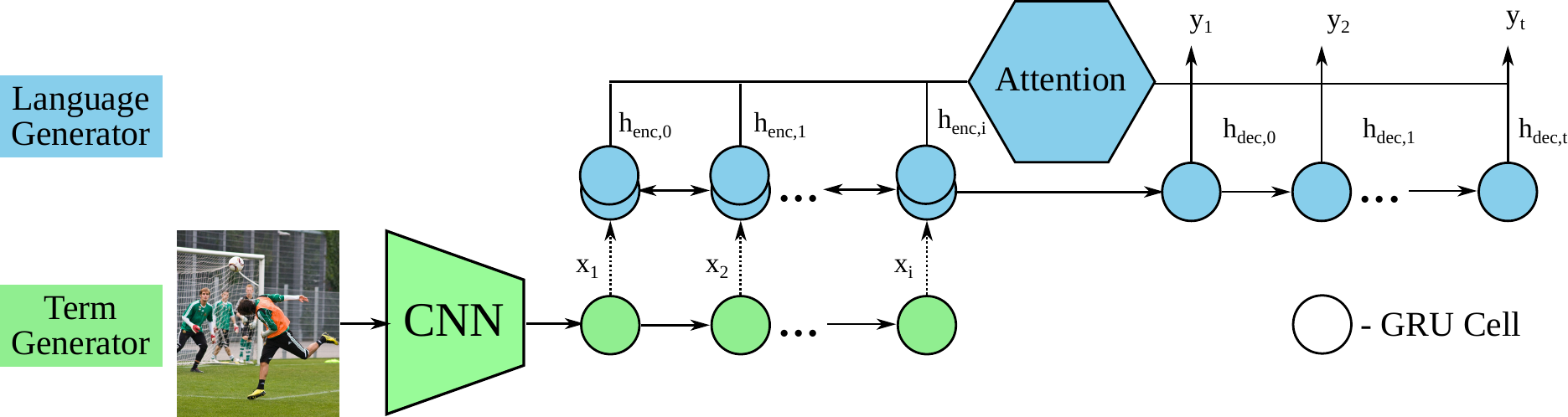}
	\caption{An overview of the {\it\mymodel} model. The \termgen network (in {\color{green} green}) is shown in the lower left. The \langgen network is in the upper right (in {\color{blue} blue}) .
	}
	\label{fig:model_full}\captionmoveup
\end{figure*}

\rev{We propose a novel encoder-decoder model for generating semantically relevant styled captions. First this model maps the image into a semantic term representation via the \termgen, then the \langgen uses these terms to generate a caption in the target style. This
is illustrated in Figure~\ref{fig:model_full}.}
    
The lower left of Figure~\ref{fig:model_full} describes the \termgen, which takes an image as input, extracts features using a CNN (Convolutional Neural Network) and then generates an ordered term sequence summarising the image semantics. The upper right of Figure~\ref{fig:model_full} describes the \langgen, which takes the term sequence as input, encodes it with an RNN (Recurrent Neural Network) and then using an attention based RNN decodes it into natural language with a specific style. We design a two-stage learning strategy enabling us to learn the \termgen network using only a standard image caption dataset, such as MSCOCO~\cite{chen2015microsoft}, and learn the \langgen network on styled text data, such as romantic novels.
The remainder of this section introduces our semantic representation and encoder-decoder neural network, while the learning method is discussed in Section~\ref{sec:learn}.

\subsection{Semantic term representation}
\label{ssec:semantic_terms}

To generate image captions that are both semantically relevant and appropriately styled, our structured semantic representation should capture visual semantics and be independent of linguistic style. We would like all semantics to be represented in the image, while language constructs that are stylistic in nature can be freely chosen by the \langgen. Our representation also needs to fully capture the semantics, to avoid teaching the \langgen to invent semantics. Since we also wish to train the \langgen without images, we need a representation that can be extracted from text alone. In Section~\ref{sec:results}, we show this semantic sequence preserves the majority of real-world image and text semantics.

Formally, given a sentence ${\bf w}=\{w_1,w_2,\cdots, w_n\}$ with $w_i\in\mathcal{V}^{in}$, we define a set of rules mapping it to our ordered semantic terms ${\bf x}=\{x_1,x_2,\cdots, x_M\}, x_i\in \mathcal{V}^{term}$. Our goal is to define \rev{a set of semantic terms and mapping} rules broad enough to encompass the semantics of both images and stylistic texts, and yet specific enough to avoid encoding style. Inspired by computational stylistics we construct three sets of rules: 

\noindent \textbf{A. Filtering non-semantic words.} Function words are known to encode style rather than semantics, and are often used in authorship identification models~\cite{Argamon2005,VanHalteren2005,Argamon2003}. Here we remove function words in order to encode semantics and strip out style. From input sentence $s$, we filter English stop-words and a small list of additional terms, either informal e.g. ``nah", the result of tokenization e.g. ``nt", or numbers e.g. ``one", ``two". Using Parts Of Speech (POS) tags we further remove: punctuation, adverbs, adjectives, pronouns and conjunctions. 
This importance ordering of POS types is derived from a data-driven perplexity evaluation described in the appendix Section~\ref{ssec:explore_semantic_terms}. 
Throughout this process we preserve common collocations such as ``hot dog'' and ``fire hydrant''. These collocations are from a pre-defined list, but automatic approaches~\cite{Wehrli2010} could also be used.

\noindent \textbf{B. Lemmatization and tagging.} Words from a sentence are converted to semantic terms to remove common surface variations. 
For most words we choose to lemmatize and concatenate with the POS tag, e.g. ``rock" becomes "rock\_NOUN". Lemmatization allows terms to be used more freely by the \langgen, enabling stylistic choices such as tense and active/passive voice. POS tags distinguish among different senses of the same word, for example the verb ``rock'' and the noun ``rock'' are disparate. We use the spaCy\footnote{\url{https://github.com/explosion/spaCy/tree/v1.9.0}} natural language toolkit for lemmatization and POS tagging. 

\noindent \textbf{C. Verb abstraction.} Verbs are replaced with a FrameNet~\cite{Baker1998} frame, preserving much of the semantics without enforcing a particular word choice. FrameNet is a lexical database of \rev{semantic frames, which are} a conceptual structure for describing events, relations, or objects along with their participants. For example, {\it sitting}, {\it laying}, {\it parking} all map to the {\it Placing} semantic frame. Table~\ref{tab:frame_examples} contains five commonly used verb frames. 
We use the semantic role labelling tool SEMAFOR~\cite{Kshirsagar2015} to annotate frames. We then map these raw frames into a reduced frame vocabulary, consisting of frames occurring over 200 times in the MSCOCO training set. Out-of-vocabulary frames are mapped to an in-vocabulary ancestor via the FrameNet hierarchy. Failing this, the frame is filtered out. Intuitively, frames not occurring frequently in the MSCOCO set, and with no frequent ancestors, are unlikely to be visually grounded -- for example the frame \textit{Certainty}, with word lemmas \textit{believe} and \textit{trust}, is a frame with no obvious visual grounding.

\begin{table}
\small
	\centering
	\begin{tabular}{|l|l|}
		\hline \textbf{Frame (count)} & \textbf{Common MSCOCO verbs} \\ \hline
		\hline Placing (86,262) & sitting, parked, laying, hanging, leaning \\ 
		\hline Posture (45,150)& standing, lying, seated, kneeling, bends \\ 
		\hline Containing (32,040)& holding, holds, held, hold \\ 
		\hline Motion (22,378)& flying, going, swinging, fly, floating \\ 
		\hline Self motion (21,118)& walking, walks, walk, swimming \\ 
		\hline 
	\end{tabular}
	\caption{The most common frames in the MSCOCO training set with frequency counts (in 596K training captions) \rev{and the most common verbs which instantiate them}.}
	\label{tab:frame_examples}\captionmoveup
\end{table}

The order of semantic terms is identical to the order in the original sentence. Results (Sec~\ref{sec:results}) show training with this ground truth ordering helps performance.

\subsection{Generating semantic terms from images}
\label{ssec:gen_terms}

We design a \termgen network that maps an input image, denoted $I$, to an ordered sequences of semantic terms ${\bf x} = \{x_1, x_2, x_i, ..., x_M\}, x_i \in \mathcal{V}^{term}$.
This is achieved with a CNN+RNN structure inspired by Show and Tell~\cite{vinyals2015show}, and illustrated in the lower left of Figure~\ref{fig:model_full}.
The image feature is extracted from the second last layer of the Inception-v3~\cite{Szegedy2016} CNN pre-trained on ImageNet~\cite{Russakovsky2015}. It passes through a densely connected layer, and then is provided as input to an RNN with Gated Recurrent Unit (GRU) cells~\cite{cho2014gru}. The term list ${\bf x}$ is shorter than a full sentence, which speeds up training and alleviates the effect of forgetting long sequences.

At each time-step $i$, there are two inputs to the GRU cell. The first is the previous hidden state ${\bf h}_{i-1}$ summarising the image $I$ and term history ${x_1,...,x_{i-1}}$, the second is the embedding vector ${\bf E}_{x_i}$ of the current term. A fully connected layer with softmax non-linearity takes the output ${\bf h}_{i}$ and produces a categorical distribution for the next term in the sequence $x_{i+1}$. Argmax decoding can be used to recover the entire term sequence from the \rev{conditional probabilities}:
\eqmoveup\begin{dmath}
{x_{i+1} = \argmax_{j \in \mathcal{V}^{term}} P(x_{i+1}=j | I, x_i...x_1)}
\end{dmath}\eqmoveup
We set $x_1$ to be a beginning-of-sequence token and terminate when the sequence exceeds a maximum length or the end-of-sequence token is generated.

\subsection{Generating styled descriptions}
\label{ssec:gen_lang}

The \langgen, shown in the upper right of Figure~\ref{fig:model_full}, maps from a list of semantic terms to a sentence
with a specific style. For example, given the term list ``\textit{dog\_NOUN}", ``\textit{Self\_motion\_FRAME}", ``\textit{grass\_NOUN}", a suitable target can be ``\textit{The dog bounded through the fresh grass.}".
Given the list of semantic terms ${\bf x}$, we generate an
output caption ${\bf y} = \{y_1, y_2, y_t, ..., y_L\}, y_t \in \mathcal{V}^{out}$ -- where $\mathcal{V}^{out}$ is the output word vocabulary. To do so, we learn
an RNN sequence-to-sequence \langgen network with attention over the input sequence, using styled text without corresponding paired images.

The encoder component for sequence ${\bf x}$ consists of a Bidirectional RNN~\cite{Schuster1997} with GRU cells and a learn-able term to vector embedding. The Bidirectional RNN is implemented as two independent RNNs running in opposite directions with shared term embeddings. Hidden outputs from the forward RNN ${\bf h}_{fwd,i}$ and the backward RNN ${\bf h}_{bak,i}$ are concatenated to form the hidden outputs of the encoder ${\bf h}_{enc,i} = [{\bf h}_{fwd,i}, {\bf h}_{bak,i}]$. The last of these hidden outputs is used to initialise the hidden state of the decoder ${\bf h}_{dec,0} = {\bf h}_{enc, M}$. The decoder itself is a unidirectional RNN (only a single forwards RNN) with GRU cells, learn-able word embeddings, attention layer, and a softmax output layer.

The attention layer connects selectively weighted encoder hidden states directly to decoder cells, using weightings defined by a learnt similarity (Equations~\ref{eq:att_learnt_sim} \& ~\ref{eq:att_context}). This avoids compressing the entire sequence into a single fixed length vector which improves performance in sequence-to-sequence modelling~\cite{Wu2016,Sutskever2014,Luong2015a}. Attention vector ${\bf a}_{t} = (a_{t,1}, ..., a_{t,i}, ..., a_{t,M})$ quantifies the importance of the input term $i$ to the current output time-step $t$. We compute the attention vector as a softmax over similarity ${\bf v}_{t}$ with learnt weight matrix $W^a$, defined as:
\eqmoveup\begin{dmath}
{v_{t, i} = {\bf h}_{enc, i}^\top W^a {\bf h}_{dec, t}} \\
{a_{t, i} = \exp(v_{t, i}) / \sum_{j=1}^{M}\exp(v_{t, j})}
\label{eq:att_learnt_sim}
\end{dmath}\eqmoveup
Using the attention we compute a context vector that summarises the important hidden outputs of the encoder for the current decoder time step. The context vector at step $t$ is defined as a weighted sum of the hidden outputs:
\eqmoveup\begin{dmath}
{{\bf c}_{t} = \sum_{i=1}^{M}a_{t,i}{\bf h}_{enc, i}}
\label{eq:att_context}
\end{dmath}\eqmoveup
To produce the output distribution we concatenate the context vector ${\bf c}_{t}$ with the hidden output of the decoder component ${\bf h}_{dec, t}$, and apply a fully connected layer with softmax non-linearity:
\eqmoveup\begin{dmath}
{{\bf h}_{out, t} = W^{out} [{\bf c}_{t}, {\bf h}_{dec, t}] + {\bf b}^{out}} \\
{p(y_t = k | {\bf x}) = \exp(h_{out, t, k}) / \sum_{j=1}^{|\mathcal{V}^{out}|}\exp(h_{out, t, j})}
\end{dmath}\eqmoveup
Here
$|\mathcal{V}^{out}|$
denotes the output vocabulary size, $[{\bf c}_{t}, {\bf h}_{dec, t}]$ denotes vector concatenation, $W^{out}, {\bf b}^{out}$ are both learnt parameter of the output layer, and $t$ is an index to the current element of the decoded sequence. 


\section{Learning with Unpaired Styled Texts}
\label{sec:learn}

The \mymodel
network learns on existing image caption datasets with only factual descriptions, plus a large set of styled texts without aligned images. 
To achieve this, we develop a two-stage training strategy for the \termgen and \langgen.

\subsection{Training the \termgenr}

We train the \termgen network on an image caption dataset with factual descriptions, such as MSCOCO. The ground truth semantic sequence for each image is constructed from the corresponding ground truth descriptive captions by following the steps in Section~\ref{ssec:semantic_terms}.

For each image, the loss function is the mean categorical cross entropy over semantic terms in the sequence:
\eqmoveup\begin{dmath}
	{\mathcal{L} = - \frac{1}{M} \sum_{i=1}^{M}{\log{p(x_i=\hat{x}_i | I, \hat{x}_{i-1} ... \hat{x}_{1})}}}
	\label{eq:cce}
\end{dmath}\eqmoveup
Here $\hat{x}$ denotes ground truth terms.
At training time the input terms $\hat{x}_{i-1} ... \hat{x}_1$ are ground truth -- this is the common teacher forcing technique~\cite{Williams1989}.  
We found that schedule sampling~\cite{Bengio2015} -- where sampled outputs are fed as inputs during training -- did not improve performance,
despite recent work on
longer sequences achieving small gains~\cite{Vinyals2017}.

\subsection{Training the \langgenr}
\label{ssec:join_learning}

The \langgen described in Section~\ref{ssec:gen_lang} takes a semantic term sequence ${\bf x}$ as input and generates a sentence ${\bf y}$ in the desired style. To create training data we take a training sentence ${\bf y}$ and map it to a semantic sequence ${\bf x}$ according to the steps in Section~\ref{ssec:semantic_terms}. The loss function is categorical cross entropy.

We train the \langgen with both styled and descriptive sentences. This produces a richer language model able to use descriptive terms infrequent in styled sentences. Training only requires text, making it adaptable to many different datasets.

Concatenating both datasets leads to two possible output styles; however, we wish to specify the style. Our solution is to provide a \styleterm during training and testing.  
Specifically, our \langgen network is trained on both the descriptive captions and the styled text with a \styleterm, indicating provenance, appended to each input sequence. As our encoder is bidirectional we expect it is not sensitive to term placement at the beginning or end of the sequence, while a term at every time step would increase model complexity. This technique has previously been used is sequence-to-sequence models for many-to-many translation~\cite{johnson2017googlemling}. In Section~\ref{sec:results} we demonstrate that purely descriptive or styled captions can be generated from a single trained model by changing the \styleterm.

\titlemoveup
\section{Evaluation settings}
\label{sec:exp_design}
\textmoveup

Both the \termgen and \langgen use separate 512 dimensional GRUs and term or word embedding vectors. The \termgen has a vocabulary of 10000 terms while the \langgen has two vocabularies: one for encoder input another for the decoder -- both vocabularies have 20000 entries to account for a broader scope. The number of intersecting terms between the \termgen and the \langgen is 8266 with both datasets, and 6736 without. Image embeddings come from the second last layer of the Inception-v3 CNN~\cite{Szegedy2016} and are 2048 dimensional.

Learning uses mini-batch stochastic gradient decent method ADAM~\cite{Kingma2015} with learning rate 0.001. We clip gradients to $[-5, 5]$ and apply dropout to image and sentence embeddings. The mini-batch size is 128 for both the \termgen and the \langgen. 
For the \langgen each mini-batch is composed of 64 styled sentences and 64 image captions. To achieve this one-to-one ratio we randomly down-sample the larger of the two datasets at the start of each epoch.

At test time both the \termgen and the \langgen use greedy decoding: the most likely word is chosen as input for the next time step. The code and trained models are released online~\footnoteref{ft:data}.

\subsection{Datasets}
\label{ssec:datasets}

\noindent{\bf Descriptive image captions} 
come from the MSCOCO dataset~\cite{chen2015microsoft} of 82783 training images and 40504 validation images, with 5 descriptive captions each.
It is common practice~\cite{vinyals2015show} to merge a large portion of this validation set into the training set to improve captioning performance. We reserve 4000 images in the validation set as a test set, the rest we merged into training set. The resulting training set has 119287 images and 596435 captions.

\noindent{\bf The styled text} consists of 1567 romance novels from bookcorpus~\cite{Zhu2015} -- comprising 596MB of text and 9.3 million sentences. We filter out sentences with less than 10 characters, less than 4 words, or more than 20 words. We further filter sentences not containing any of the 300 most frequent non stop-words from the MSCOCO dataset -- leaving 2.5 million sentences that are more likely to be relevant for captioning images. 
Our stop-word list is from NLTK~\cite{Bird2009} and comparisons are on stemmed words. For faster training and to balance the styled and descriptive datasets we further down-sample to 578,717 sentences, with preference given to sentences containing the most frequent MSCOCO words. We remove all but the most basic punctuation (commas, full stops and apostrophes), convert to lower-case, tokenise and replace numbers with a special token.

The StyleNet~\cite{gan2017stylenet} test set was not released publicly at the time of writing, \rev{so we could not use it for comparisons.}

\subsection{Compared approaches}
\label{ssec:baselines}

We evaluate 6 state-of-the-art baselines and 7 variants of \textit{\mymodel}~-- extended details are in the appendix Section~\ref{sec:baselines}. 

\textbf{CNN+RNN-coco} is based on the descriptive Show+Tell model~\cite{vinyals2015show}, 
\textbf{TermRetrieval} uses the \termgen to generate a list of words, which are then used to retrieve sentences from the styled text corpus. 
\textbf{StyleNet} generates styled captions, while \textbf{StyleNet-coco} generates descriptive captions. Both are a re-implementation of the Gan~\etal~\cite{gan2017stylenet} model with minor modifications to ensure convergence on our dataset.
\textbf{neural-storyteller} is a model trained on romance text (from the same source as ours) and released by Kiros~\cite{Kiros2015b}. 
\textbf{JointEmbedding} maps images and sentences to a continuous multi-modal vector space~\cite{Kiros2014a}, and uses a separate decoder, trained on the romance text, to decode from this space. 

\textbf{\mymodel} is our full model. \textbf{\mymodel-unordered} is a variant of \textit{\mymodel} with a randomised semantic term ordering; 
\textbf{\mymodel-words} is a variant where the semantic terms are raw words -- they are not POS tagged, lemmatized or mapped to FrameNet frames;
\textbf{\mymodel-lempos} is a variant where the semantic terms are lemmatized and POS tagged, but verbs are not mapped to FrameNet frames;
\textbf{\mymodel-romonly} is \textit{\mymodel} with the language generator trained only on the romantic novel dataset.
\textbf{\mymodel-cocoonly} is \textit{\mymodel} trained only on the MSCOCO dataset.
\textbf{\mymodel-coco} is \textit{\mymodel} trained on both datasets but with a MSCOCO \styleterm used at test time to indicate descriptive captions should be generated.

\subsection{Evaluation metrics}
\label{ssec:metrics}

\noindent{\bf Automatic \sout{content} relevance metrics.}

Widely-used captioning metrics such as BLEU~\cite{Papineni2002}, METEOR~\cite{Denkowski2014} and CIDEr~\cite{Vedantam2015} are based on n-gram overlap. They are less relevant to stylised captioning since the goal is to change wording while preserving semantics. We include them for descriptive captions (Table~\ref{tab:coco_cap_res}); results for stylised captions are in the supplement~\cite{suppliment}.
The SPICE~\cite{Anderson2016} metric computes an f-score over semantic tuples extracted from MSCOCO reference sentences~\cite{chen2015microsoft}. This is less dependent on exact n-gram overlap, and is strongly correlated with human judgements of descriptiveness. \rev{In the following we interchangably use the terms descriptiveness and relevance.}

\noindent{\bf Automatic style metrics.} 
To the best of our knowledge, there are no well-recognised measures for style adherence.
We propose three metrics, the first two use a language model in the target style, 
the second is a high-accuracy style classifier.
LM is our first language model metric, it is the average perplexity in bits per word under a 4-gram model~\cite{Heafield2013} built on the romance novels. Lower scores indicate stronger style. The GRULM metric is the bits per word under a GRU language model, with the structure of the \langgen decoder without attention.
The CLassifier Fraction (CLF) metric, is the fraction of generated captions classified as styled by a binary classifier.
This classifier is logistic regression with 1,2-gram occurrence features trained on styled sentences and MSCOCO training captions. Its cross-validation precision is 0.992 at a recall of 0.991. We have released all three models~\footnoteref{ft:data}.

\noindent{\bf Human evaluations of \rev{relevance} and style.}

Automatic evaluation does not give a full picture of a captioning systems performance~\cite{chen2015microsoft}; 
human evaluation can help us to better understand its strengths and weaknesses, 
with the end user in mind. We evaluate each image-caption pair 
with two crowd-sourced tasks on the CrowdFlower\footnote{\url{https://www.crowdflower.com}} platform. 
The first measures how descriptive a caption is to an image on a four point scale -- from unrelated (1) to clear and accurate (4). 
The second task evaluates the degree of style transfer. 
We ask the evaluator to choose among three mutually exclusive options -- that the caption: is likely to be part of a story related to the image ({\it story}), is from someone trying to describe the image to you ({\it desc}), or is completely unrelated to the image ({\it unrelated}). 
Note that most sentences
in a romance novel are not identifiably romantic once taken out of context. 
Being part of a story is an identifiable property for a single sentence. 
We choose this over shareability, as used by Gan~\etal~\cite{gan2017stylenet}, 
since being part of a story more concisely captures the literary quality of the styled text.
We separate the descriptiveness and story aspects of human evaluation,
after pilot runs found that the answer to descriptiveness
 interferes with the judgement about being part of a story.
Using each method we caption the same 300 random test images, and evaluate each with 3 workers -- a total of 900 judgements per method. More details on the crowd-sourced evaluation, including a complete list of questions and guideline text, can be found in the appendix Section~\ref{sec:human}.

\titlemoveup
\section{Results}
\label{sec:results}
\textmoveup

\begin{figure}
	\begin{minipage}[b][][b]{0.23\textwidth}
	\subfigure[]{
	  \centering
	  \includegraphics[width=\textwidth]{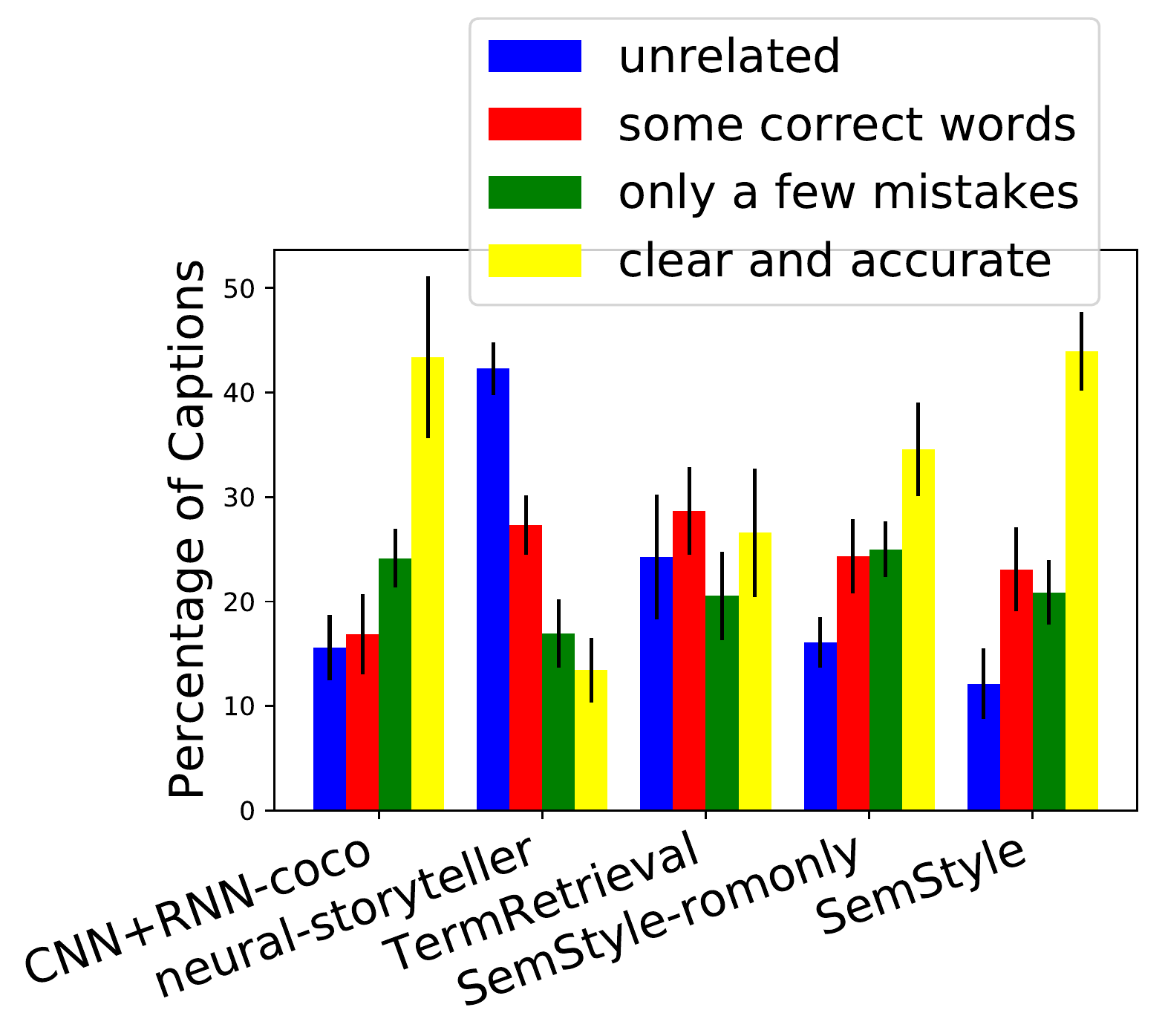}
	  \label{fig:human_eval_desc}
	}
	\end{minipage}
	\begin{minipage}[b][][b]{0.23\textwidth}
	\subfigure[]{
	  \centering
	  \includegraphics[width=\textwidth]{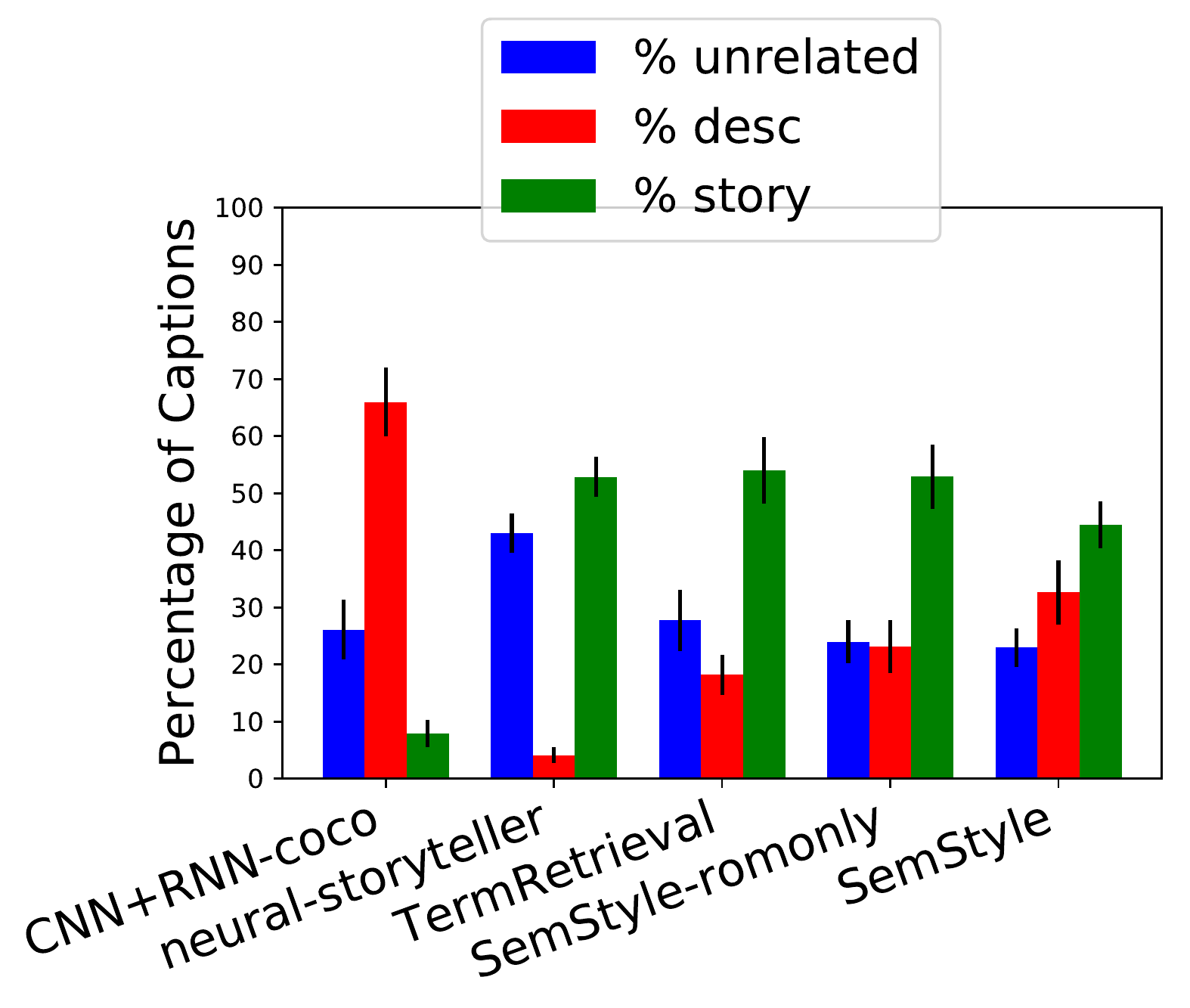}
	  \label{fig:human_eval_story}
	}
	\end{minipage}
	\caption{Human evaluations for \mymodel and selected baselines. \protect\subref{fig:human_eval_desc} relevance measured on a four point scale, reported as percentage of generated captions at each level with 0.95 confidence interval error bars. \protect\subref{fig:human_eval_story} style conformity as a percentage of captions: unrelated to the image content, a basic description of the image, or part of a story relating to the image.}
	\label{fig:human_eval}\captionmoveup
\end{figure}

Table~\ref{tab:coco_cap_res} summarizes measurements of content relevance against factual (MSCOCO) captions. Table~\ref{tab:style_cap_res} and Figure~\ref{fig:human_eval} report automatic and human evaluations on caption style learned from romance novels.


\begin{table*}
\begin{center}
\begin{tabular}{|l||c|c|c|c|c||c|c|c|}
\hline \textit{Model} & \textit{BLEU-1} & \textit{BLEU-4} & \textit{METEOR} & \textit{CIDEr} & \textit{SPICE} & \textit{CLF} & \textit{LM} & \textit{GRULM}\\ \hline
\hline CNN+RNN-coco & 0.667 & 0.238 & 0.224 & 0.772 & 0.154 & 0.001 & 6.591 & 6.270\\ 
\hline StyleNet-coco & 0.643 & 0.212 & 0.205 & 0.664 & 0.135 & 0.0 & 6.349 & 5.977\\
\hline \mymodel-cocoonly & 0.651 & 0.235 & 0.218 & 0.764 & 0.159 & 0.002 & 6.876 & 6.507\\
\hline \mymodel-coco & 0.653 & 0.238 & 0.219 & 0.769 & 0.157 & 0.003 & 6.905 & 6.691\\ \hline
\end{tabular}
\caption{Evaluating caption relevance 
on the MSCOCO dataset. For metrics see Sec.~\ref{ssec:metrics}, for approaches see Sec.~\ref{ssec:baselines}.}
\label{tab:coco_cap_res}
\end{center}\captionmoveup\captionmoveup
\end{table*}

\noindent{\bf Evaluating \sout{content} relevance.}
\textit{\mymodel-coco} \rev{generates descriptive captions because the descriptive \styleterm is used.} It achieves semantic relevance scores comparable to the 
\textit{CNN+RNN-coco}, with a SPICE of 0.157 vs 0.154, and BLEU-4 of 0.238 for both. 
This demonstrates that using semantic terms is a competitive way to distil image semantics, 
and that the \termgen and \langgen constitute an effective vision-to-language pipeline. \rev{Moreover, {\it \mymodel} can be configured to generate different caption styles just by changing the \styleterm at test time -- the complement of the CLF metric shows \textit{\mymodel-coco} captions are classified as descriptive in 99.7\% of cases.}

\noindent{\bf Evaluating style.}
\textit{\mymodel} succeeds in generating styled captions in 58.9\% of cases, as judged by CLF, and receives a SPICE score of 0.144. The baselines {\it TermRetrieval}, \textit{neural-storyteller} and {\it JointEmbedding} have significantly higher CLF scores, but much lower SPICE scores. \textit{TermRetrieval} produces weakly descriptive sentences (SPICE of 0.088) because it is limited to reproducing the exact text of the styled dataset which yields lower recall for image semantics. Both \textit{neural-storyteller} (SPICE 0.057), and \textit{JointEmbedding} (SPICE 0.046), \rev{decode from a single embedding vector allowing less control over semantics than \mymodel. This leads to weaker caption relevance.}
\textit{StyleNet-coco} produces factual sentences with comparable BLEU and SPICE scores. 
However, \textit{StyleNet} 
produces styled sentences less frequently (CLF 41.5\%) and with significantly lower 
semantic relevance -- SPICE of 0.010 compared to 0.144 for {\it\mymodel}.
We observe that the original {\it StyleNet} dataset~\cite{gan2017stylenet}
mostly consists of factual captions re-written by adding or editing a few words.
The romance novels in the book corpus, on the other hand, 
have very different linguistic patterns to COCO captions. 
We posit that the factored input weights in {\it StyleNet} work well for 
small edits, but have difficulty capturing richer and more drastic changes.
For {\it \mymodel}, the semantic term space and a separate \langgen make it amenable to larger stylistic changes.

\noindent{\bf Coverage of semantic terms.} We find that most of the terms generated by the \termgen are represented in the final caption by the \langgen. Of the Non-FrameNet terms 94\% are represented, while 96\% of \rev{verb frames} are represented. Evaluating the full \textit{\mymodel} pipeline involves mapping generated and ground truth captions to our semantic term space and then calculating multi-reference precision (BLEU-1) and recall (ROUGE-1). \textit{\mymodel} gets a higher precision 0.626 and recall 0.517 than all styled caption baselines, and is close to the best descriptive model \textit{\mymodel-cocoonly} with precision 0.636 and recall 0.531. Full results are in the appendix Section~\ref{ssec:pr_term_space}.


\begin{table}
	\centering
	\small
	\begin{tabular}{|l||c|c|c|c|}
	\hline \textit{Model} & \textit{SPICE} & \textit{CLF} & \textit{LM} & \textit{GRULM}\\ \hline
	\hline StyleNet & 0.010 & 0.415 & 7.487 & 6.830\\
	\hline TermRetrieval & 0.088 & 0.945 & 3.758 & 4.438\\   
	\hline neural-storyteller & 0.057 & 0.983 & 5.349 & 5.342\\ 
	\hline JointEmbedding & 0.046 & 0.99 & 3.978 & 3.790\\ \hline
	\hline \mymodel-unordered & 0.134 & 0.501 & 5.560 & 5.201\\ 
	\hline \mymodel-words & 0.146 & 0.407 & 5.208 & 5.096\\ 
	\hline \mymodel-lempos & 0.148 & 0.533 & 5.240 & 5.090\\ 
	\hline \mymodel-romonly & 0.138 & 0.770 & 4.853 & 4.699\\ 
	\hline \mymodel & 0.144 & 0.589 & 4.937 & 4.759\\ 
	\hline  
	\end{tabular}
	\caption{Evaluating styled captions with automated metrics. For \textit{SPICE} and \textit{CLF} larger is better, for \textit{LM} and \textit{GRULM} smaller is better. For metrics and baselines see Sec.~\ref{ssec:metrics} and Sec.~\ref{ssec:baselines}.}
	\label{tab:style_cap_res}\captionmoveup
\end{table}

\begin{figure*}
\centering
\includegraphics[width=0.95\textwidth]{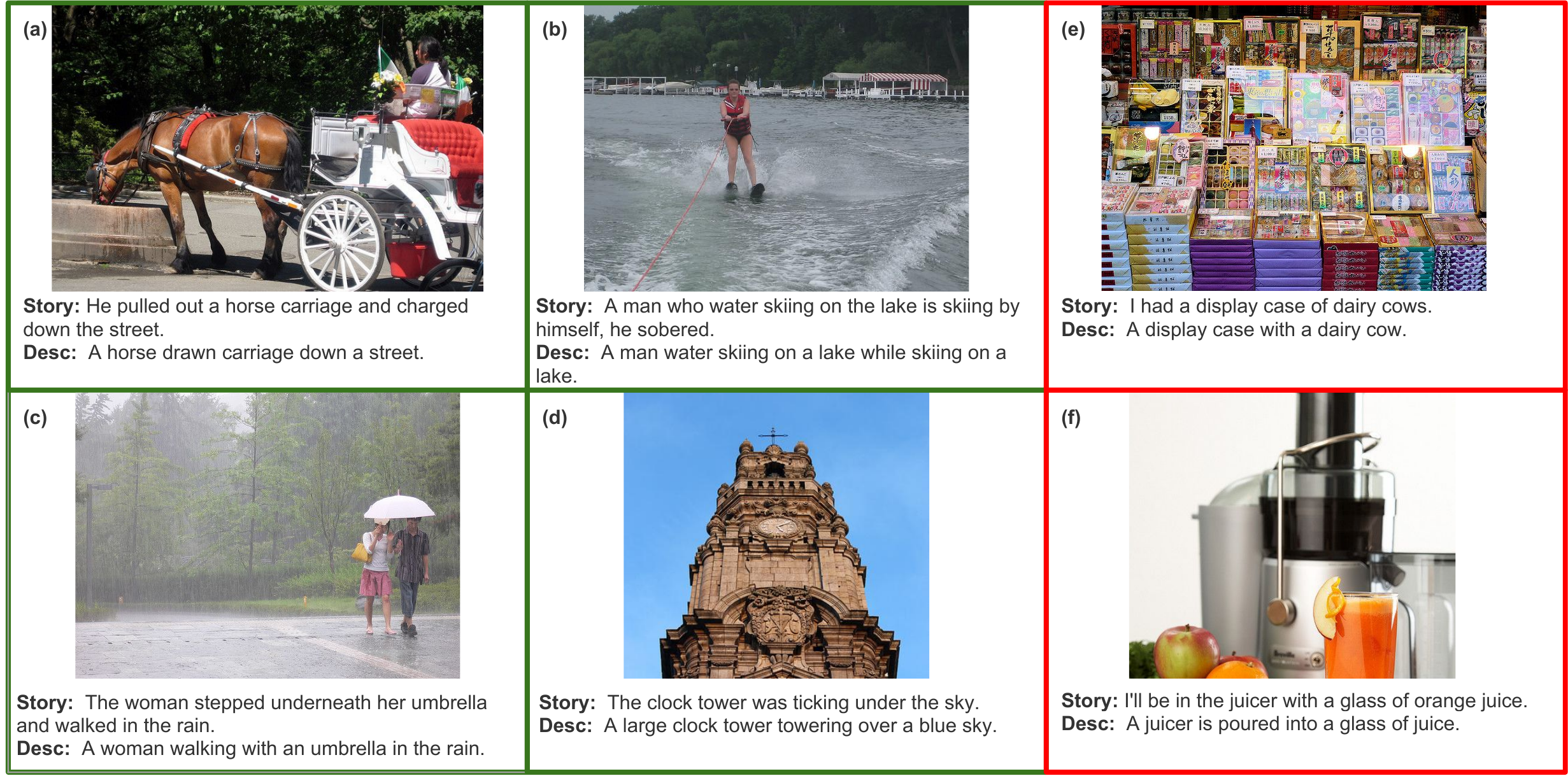}
\caption{Example results, includes styled (\textbf{Story}) output from \textit{\mymodel} and descriptive (\textbf{Desc}) output from \textit{\mymodel-coco}. Four success cases are on the left (a,b,c,d), and two failures are on the right (e,f).\vspace{-5.0mm}}
\label{fig:example_results}
\end{figure*}\textmoveup

\noindent{\bf Human evaluations}
are summarised in Figure~\ref{fig:human_eval}, tabular results and statistical significance testing are in the appendix Section~\ref{ssec:hypotest_human}.
Figure~\ref{fig:human_eval_desc} shows image-caption relevance judged on a scale of 1 (unrelated) to 4 (clear and accurate). 
\textit{StyleNet} was not included in the human evaluations since it scored significantly worse than others in the automatic metrics, especially SPICE and LM. 
{\it\mymodel} has a mean relevance of 2.97 while \textit{CNN+RNN-coco} has 2.95. In addition, only 12.2\% of {\it\mymodel} captions are judged as {\it unrelated}, the lowest among all approaches. {\it\mymodel} produces {\it clear and accurate} captions 43.8\% of the time, while \textit{CNN+RNN-coco} produces them 43.4\% of the time -- significantly higher than other approaches. \rev{As the CNN+RNN architecture is the basis of the \termgen, this indicates our semantic term mapping and separate styled \langgen do not reduce the relevance of the generated captions.}
\textit{TermRetrieval} has mean relevance 2.50, and \textit{neural-storyteller} 2.02 -- both significantly lower than \textit{\mymodel}. \textit{neural-storyteller} generates a large fraction of completely unrelated captions (42.3\%) while \textit{TermRetrieval} avoids doing so (24.4\%). \textit{\mymodel-romonly} produces fewer clear and accurate captions than \textit{\mymodel} (34.7\% vs 43.8\%), which demonstrates improved caption relevance when both training datasets are combined.

Figure~\ref{fig:human_eval_story} summarises crowd-worker choices among being {\it story-like}, {\it descriptive}, or {\it unrelated}. The two \textit{\mymodel} variants have the lowest ($<25\%$) fraction of captions that are judged {\it unrelated}.
\textit{\mymodel} generates story like captions 41.9\% of the time, which is far more frequently than the \textit{CNN+RNN-coco} trained on MSCOCO at 6.2\%. \textit{neural-storyteller} produces captions that are judged as story like 52.6\% of the time, but at the expense of 44.2\% completely unrelated captions. \textit{TermRetrieval} produces captions that are story like 55.5\% of the time and unrelated only 26.0\% of the time; however, as shown in Figure~\ref{fig:human_eval_desc}, the relevance to images is low. 

We calculate the correlation between the three new style metrics (in Section~\ref{ssec:metrics}) and human {\it story-like} judgements. Following the method of Anderson~\etal~\cite{Anderson2016}, Kendall's $\tau$ correlation co-efficient is: 0.434 for CLF, 0.150 for LM, and 0.091 for GRULM.

\noindent{\bf Evaluating modeling choices of \mymodel.} The last 5 rows of Table~\ref{tab:style_cap_res} highlight trade-offs among variants of {\it\mymodel}. Randomly ordering the semantic terms during training and testing, \textit{\mymodel-unordered}, leads to captions with less semantic relevance, shown by a SPICE of 0.134 compared to 0.144 for the full model. They also conform less to the target style with a CLF of 0.501 compared to 0.589. 

Using a raw word term space \textit{\mymodel-words} (without FrameNet, lemmatization or POS tags) gives similar semantic relevance, SPICE of 0.146 to the full models 0.144, but less styling with CLF at 0.407. Using verb lemmas rather than FrameNet terms as in \textit{\mymodel-lempos}, has a similar effect, with a slight increase in SPICE to 0.148 and a decrease in style to a CLF of 0.533. This clearly demonstrates the three components FrameNet, lemmatization and POS tags all contribute to remove style from the intermediate representation, and thus lead to output in the target style.

Learning from both datasets improves caption relevance. If we only train on the romantic novel corpus as in \textit{\mymodel-romonly}, we find strong conformity to the target style (CLF 0.770) but less semantic relevance, SPICE 0.138. Without the joint training some semantics terms from the MSCOCO dataset are never seen by the language generator at training time -- meaning their semantic content is inaccessible at test time. Our joint training approach avoids these issues and allows style selection at test time.

\noindent{\bf Example captions.} Figure~\ref{fig:example_results} shows four success cases on the left (a,b,c,d) and two failures on the right (e,f). The success cases are story-like, such as ``The woman stepped underneath her umbrella and walked in the rain." rather than ``A woman walking with an umbrella in the rain.". They also tend to use more interesting
verbs (due to FrameNet) -- ``He pulled out a horse carriage and \textbf{charged} down the street". We note \textit{\mymodel} examples use more past tense (c), definite articles (b,d), and first person view (e,f) -- statistics are included in the appendix Section~\ref{ssec:attributes_genstyle}. The failures are caused by the \termgen incorrectly identifying cows in the image (top row), or word use (``juicer'') by the \langgen that is grammatically correct, but contradicts common-sense (bottom row).

\titlemoveup
\section{Conclusion}
\textmoveup

We propose {\it\mymodel}, a method to learn visually grounded style generation from texts without paired images.
We develop a novel semantic term representation to disentangle
content and style in descriptions. This allows us to learn a mapping from an image to a sequence of semantic terms that preserves visual content, and a decoding scheme that generates a styled description. 
Future work includes learning from a richer set of styles, and developing a recognised set of automated and subjective metrics for styled captions.

\noindent{\bf Acknowledgments} 
This work is supported, in part, by the Australian Research Council via project DP180101985.
The Tesla K40 used for this research was donated by the NVIDIA Corporation. We thank S. Mishra, S. Toyer, A. Tran, S. Wu, and A. Beveridge for helpful suggestions.

{\small
\bibliographystyle{ieee}
\bibliography{ms}
}

\FloatBarrier
\newpage

\onecolumn

\section{Appendix}
\subsection{Baseline Implementation Details}
\label{sec:baselines}

Our evaluations includes 5 state-of-the-art baselines.

\textbf{CNN+RNN-coco} is based on the Show+Tell model~\cite{vinyals2015show} and trained on only the MSCOCO dataset. We use a GRU cell in place of an LSTM cell for a fairer comparison with our model. In fact, this baseline is just the \termgen component of \mymodel trained to output full sentences rather than sequences of terms. All hyper-parameter settings are the same as for the \termgen.

\textbf{TermRetrieval} uses the \termgen to generate a list of terms -- in this case the term vocabulary is words rather than lemmas with POS tags. These terms are used in an OR query of the Romance text corpus and scored with BM25~\cite{Jones2000} using hyper-parameters $b = 0.75, k_1 = 1.2$. Our query engine is Whoosh\footnote{\url{https://pypi.python.org/pypi/Whoosh/}}, which includes a tokenizer, lower-case filter, and porter stem filter. This model cannot generate caption that are not part of the romance text corpus and the same set of terms always gives the same sentence -- ie it is deterministic and only dependent on terms.

\textbf{StyleNet} is our re-implemented of the method proposed by Gan~\etal~\cite{gan2017stylenet} -- the original code was not released at the time of writing. We train it on the MSCOCO dataset and the Romantic text dataset. Our implementation follows Gan~\etal~\cite{gan2017stylenet} 
with the following implementation choices to ensure a fair comparison with other baselines.
Rather than ResNet152~\cite{He2016} features we use Inception-v3~\cite{Szegedy2016} features and a batch size of 128 for both datasets. 
When training on styled text \textit{StyleNet} requires random input noise from some unspecified distribution, we tried a few variations and found Gaussian noise with $\mu=0$ and $\sigma=0.01$ worked reasonably well. Gan~\etal suggested a training scheme where the training set alternates between descriptive and styled at the end of every epoch. We found this fails to converge, perhaps because our datasets are larger and more diverse compared with the \textit{FlickrStyle10k} dataset used in the original implementation. \textit{FlickrStyle10k}, which is not publicly released at the time of writing, contains styled captions rather than sentences sampled from novels. To ensure \textit{StyleNet} converges on our dataset we alternate between the MSCOCO dataset and Romantic text dataset after every mini-batch -- a strategy suggested by Luong~\etal~\cite{Luong2016} for multi-task sequence-to-sequence learning.

\begin{figure}
\includegraphics[width=\linewidth]{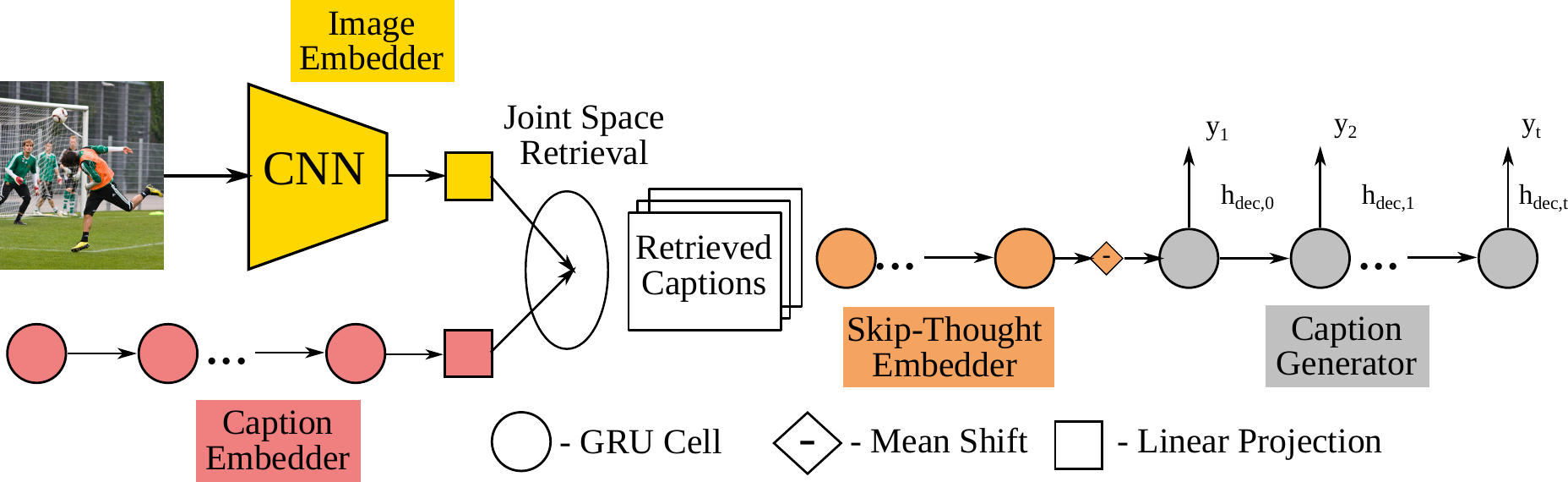}
\caption{The neural-storyteller model~\cite{Kiros2015b}, for generating short styled stories about images. The mean shift block subtracts off the mean skip-though vector for captions and adds on the mean skip-thought vector for the target style.}
\label{fig:neural_storyteller_model}
\end{figure}

\textbf{neural-storyteller} consists of pre-trained models released by Kiros~\cite{Kiros2015b} for generating styled image captions -- see Figure~\ref{fig:neural_storyteller_model}. This model, first retrieves descriptive captions using an multi-modal space~\cite{Kiros2014a} trained on MSCOCO with a VGG-19~\cite{Simonyan2015} CNN image encoder and a GRU caption encoder. Retrieved captions are encoded into skip-thought vectors~\cite{Kiros2015a}, averaged, and then style shifted. This style shift is performed by subtracting off the mean skip-though vector for captions and adding the mean skip-thought vector of text in the target style. The style shifted vector is decoded by a conditional RNN language model trained on text in the target style. The skip-though vectors are trained on the entirety of bookcorpus~\cite{Zhu2015}, while the skip-thought vector decoder is trained on the romance genre subset of bookcorpus (the same subset we have used for our models). \textit{neural-storyteller} generates passages by repeatedly sampling the decoder, we use only the first sentence because long passages would be disadvantaged by the evaluation criteria.

\begin{figure*}
	\centering
	\includegraphics[width=0.9\textwidth]{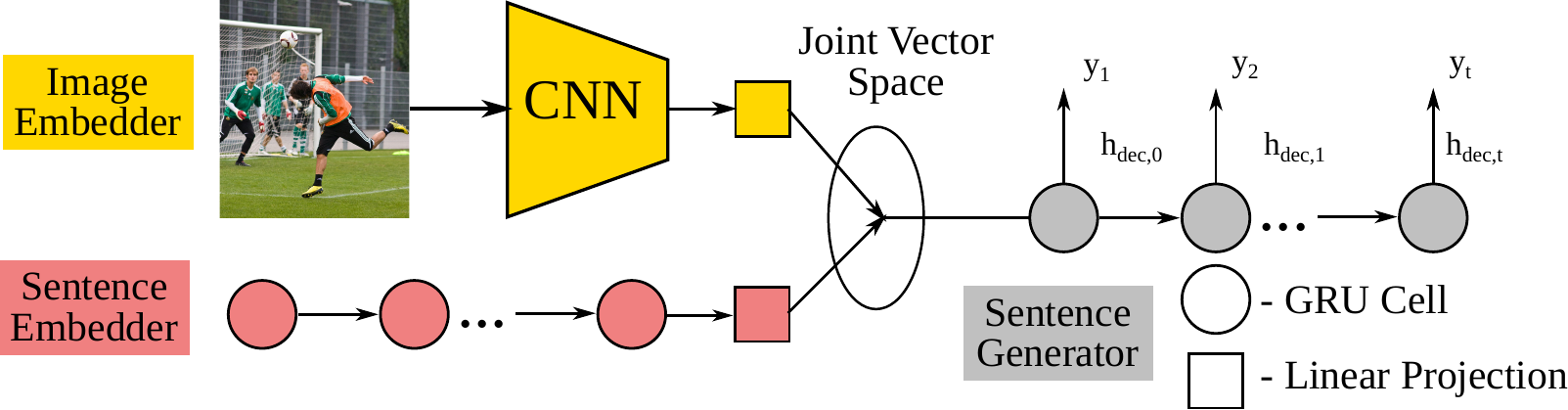}
	\caption{An overview of the {\it JointEmbedding} model. The two embedding components \imemb (in {\color{myyellow} yellow}) and \sentemb (in {\color{red} red}) are shown on the left while the \sentgen (in {\color{gray} grey}) is on the right.
	}
	\label{fig:semstyle_joint_emb}
\end{figure*}

\textbf{JointEmbedding}, shown in Figure~\ref{fig:semstyle_joint_emb}, uses a learnt multi-modal vector space as the intermediate representation. The \imemb is a projection of pre-trained Inception-v3~\cite{Szegedy2016} features $h_I$, while the \sentemb is a projection of the last hidden state of an RNN with GRU units $h_{enc}$. Formally the projections are:
\begin{dmath*}
{v_I = \tanh(W_I . h_I)}\\
{v_s = \tanh(W_s . h_{enc})}
\end{dmath*}
Denoting the projections as, $v_I$ for images and $v_s$ for captions, and the learnt projection weights as $W_I$ for images and $W_s$ for captions.
Agreement between image and caption embedding is defined as the cosine similarity:
\begin{dmath*}
g(v_I, v_s) = \frac{v_I.v_s}{|v_I||v_s|}
\end{dmath*}
To construct the space we use a noise contrastive pair-wise ranking loss suggested by Kiros et al~\cite{Kiros2014a}. Intuitively, this loss function encourages greater similarity between embeddings for paired image-captions than for un-paired images and captions.
\begin{dmath*}
\mathcal{L} = \max(0, m - g(v_I, v_s) + g(v_{I'}, v_s)) + \max(0, m - g(v_I, v_s) + g(v_I, v_{s'}))
\label{eq:semstyle_joint_emb}
\end{dmath*}
Where $s$ is the input caption pared with image $I$, while $s'$ is a randomly sampled noise contrastive caption and $I'$ the noise contrastive image. The margin $m$ is fixed to 0.1 in our experiments. 

The \sentgen is an RNN with GRU units, that decodes from the joint vector space. The loss function is categorical cross entropy given in Equation~\ref{eq:semstyle_cce}.

\begin{dmath}
	{\mathcal{L} = - \frac{1}{M} \sum_{i=1}^{M} \sum_{j\in V^m}^{}{\log{p(y_i=j | I, y_{i-1} ... y_{1})^{\mathbb{I}[y_i = j]}}}}
	\label{eq:semstyle_cce}
\end{dmath}

Training is a two stage process, first we define the joint space by learning the \imemb and the \sentemb on MSCOCO caption-image pairs. From here on the parameters of \imemb and the \sentemb are fixed.
The \sentgen is learnt separately by embedding styled sentences from the romantic novel dataset with the \sentemb into the multi-modal space and then attempting to recover the original sentence. This model has not been published previously, but is based on existing techniques for descriptive captioning~\cite{Kiros2014a}. 

\subsection{Model Variants}
\label{ssec:semstyle_model_variants}
Our full model is denoted \textit{\mymodel}. We use the following variants to assess several modelling choices.

\textbf{\mymodel-coco} is the \textit{\mymodel} model trained jointly on MSCOCO and the romance corpus with dataset indicator set to MSCOCO at test time. The output of this model should be purely descriptive.

\textbf{\mymodel-cocoonly} is the \textit{\mymodel} model trained only on MSCOCO. The output of this model should be purely descriptive.

\textbf{\mymodel-unordered} is a variant of \textit{\mymodel} with a randomised semantic term ordering. This model helps us to quantify the effect of ordering in the term space.
 
\textbf{\mymodel-words} is a variant where the semantic terms are raw words -- they are not POS tagged, lemmatized or mapped to FrameNet frames.

\textbf{\mymodel-lempos} is a variant where the semantic terms are lemmatized and POS tagged, but verbs are not mapped to FrameNet frames. This helps us to quantify the degree to which verb abstraction effects the model performance.

\textbf{\mymodel-romonly} is \textit{\mymodel} without joint training -- the language generator was trained only on the romantic novel dataset. This model helps to quantify the effect of joint training.

\subsection{Human Evaluation}
\label{sec:human}

\subsubsection{Crowd-sourcing Task Setup}

\begin{figure}
\centering
\includegraphics[width=\textwidth]{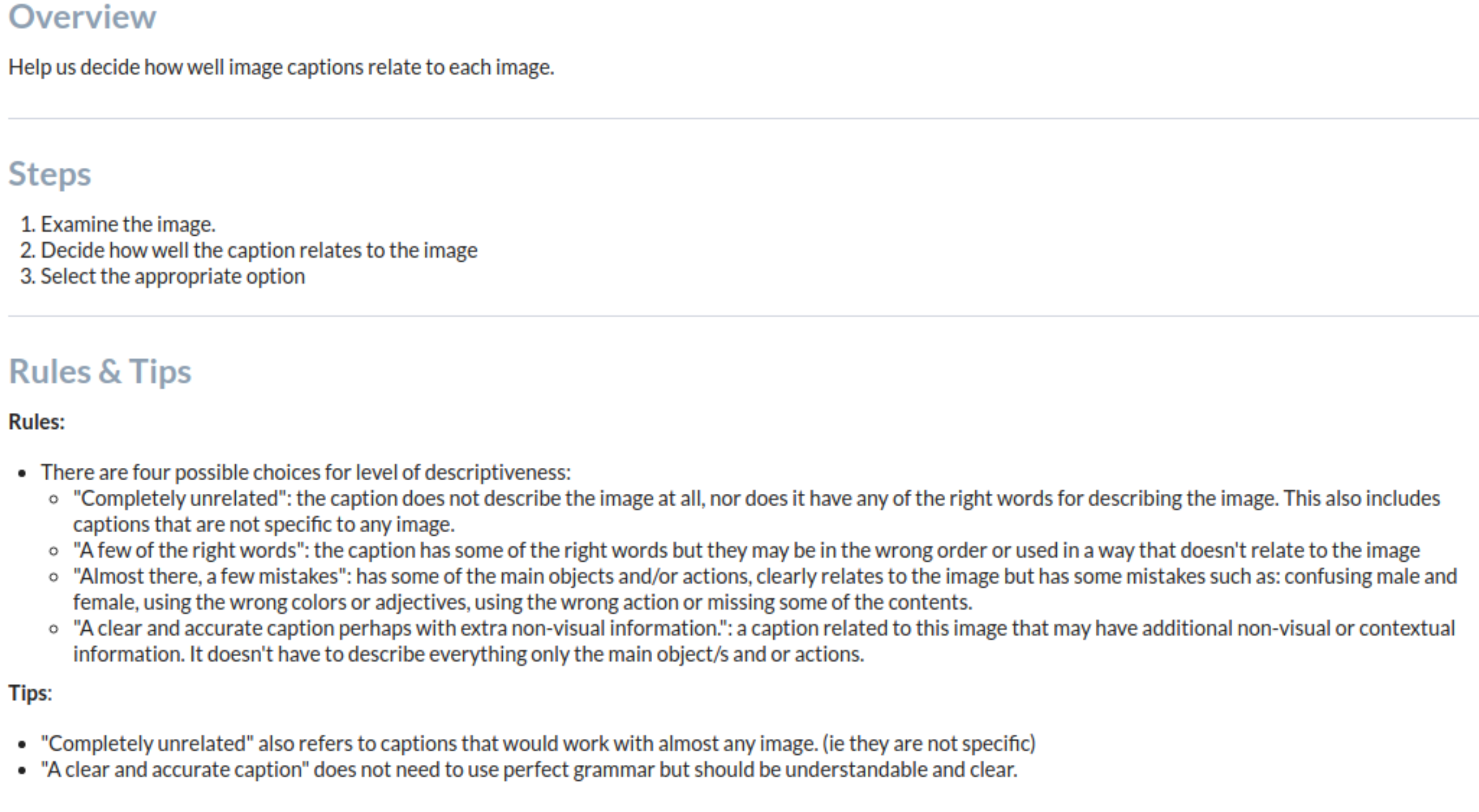}
\caption{A screen-shot of the instructions provided to workers evaluating the relevance of a caption to an image.}
\label{fig:heval_desc_inst}
\end{figure}
\begin{figure}
\centering
\includegraphics[width=\textwidth]{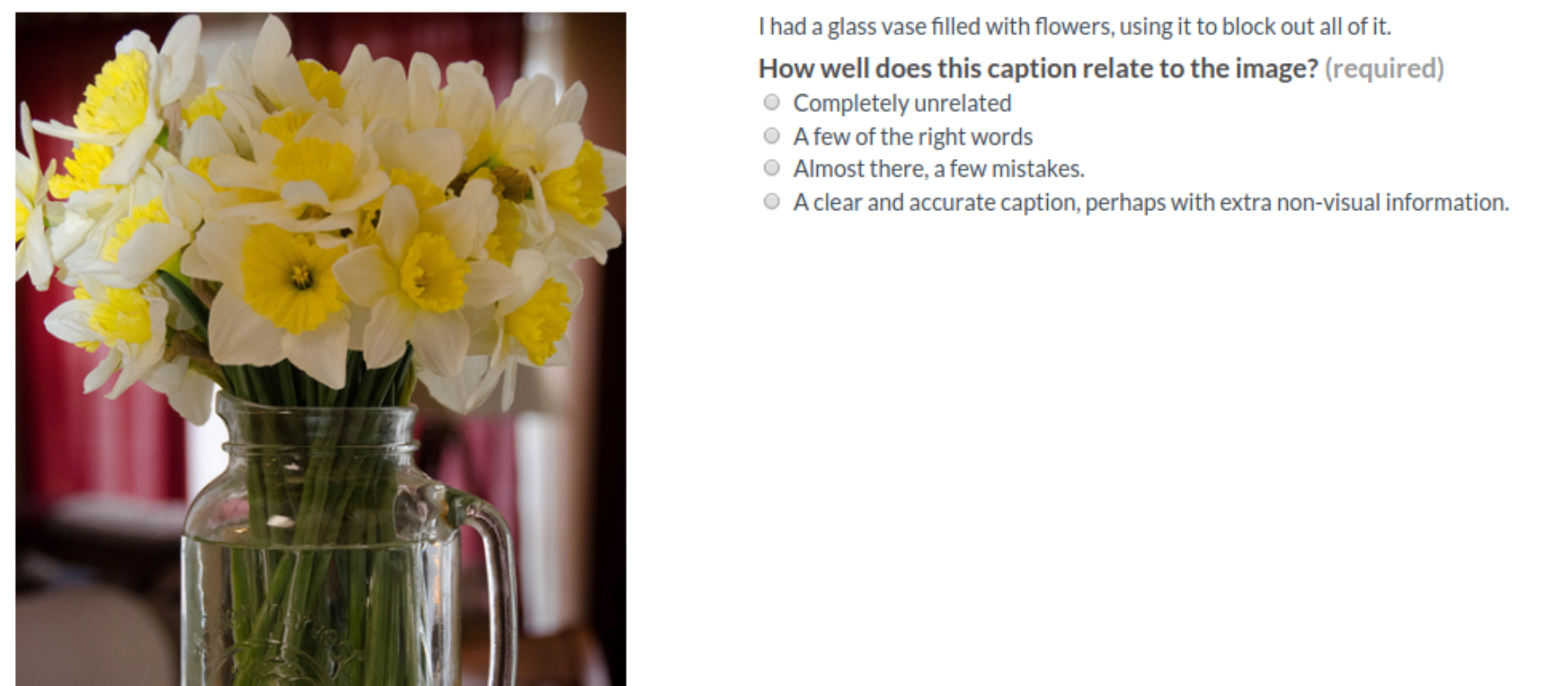}
\caption{A screen-shot of a single question asked of workers in the relevance evaluation task.}
\label{fig:heval_desc_example}
\end{figure}
\begin{figure}
\centering
\includegraphics[width=\textwidth]{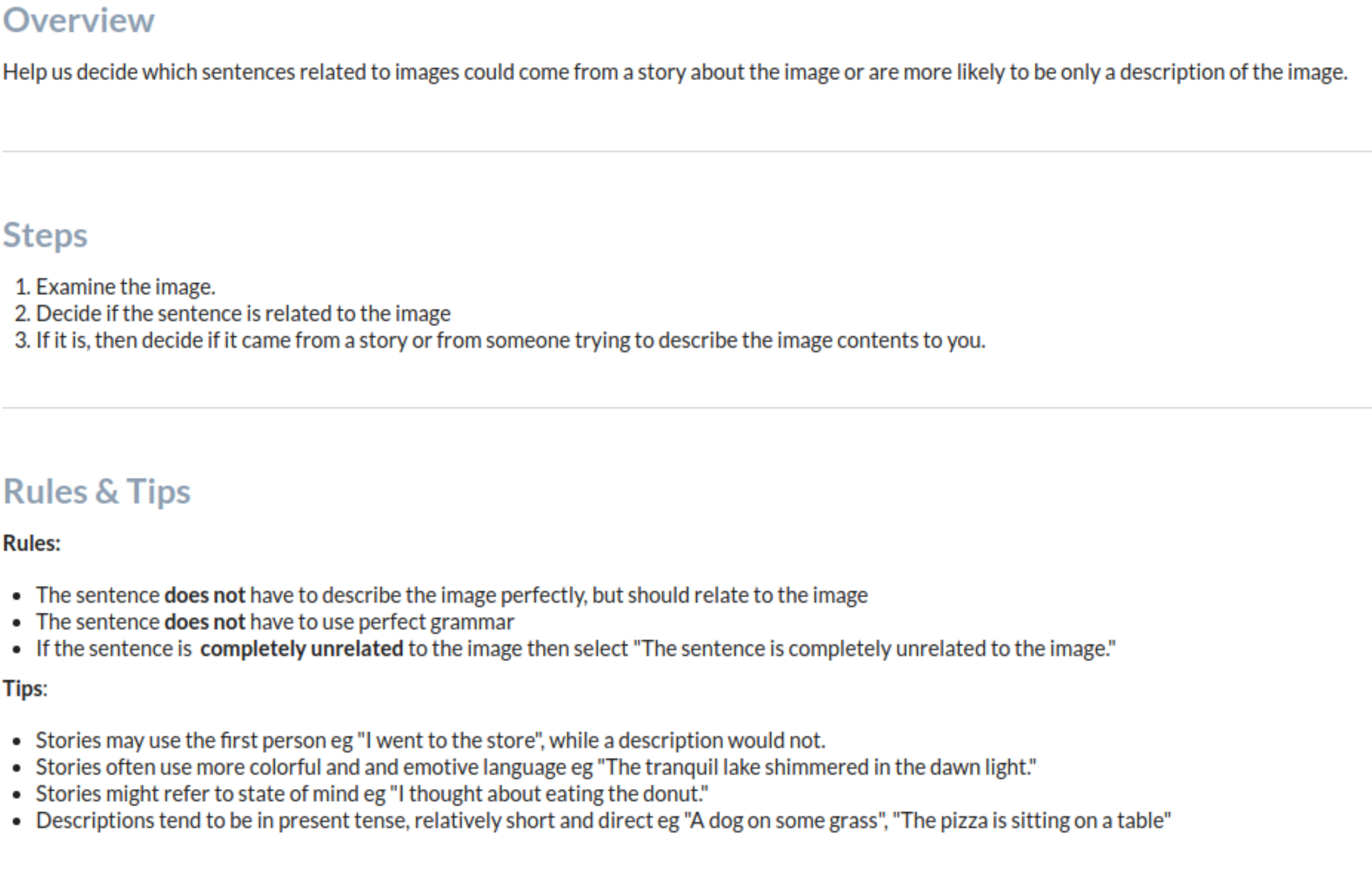}
\caption{A screen-shot of the instructions provided to workers evaluating how well a caption conforms to the desired style.}
\label{fig:heval_story_inst}
\end{figure}
\begin{figure}
\centering
\includegraphics[width=\textwidth]{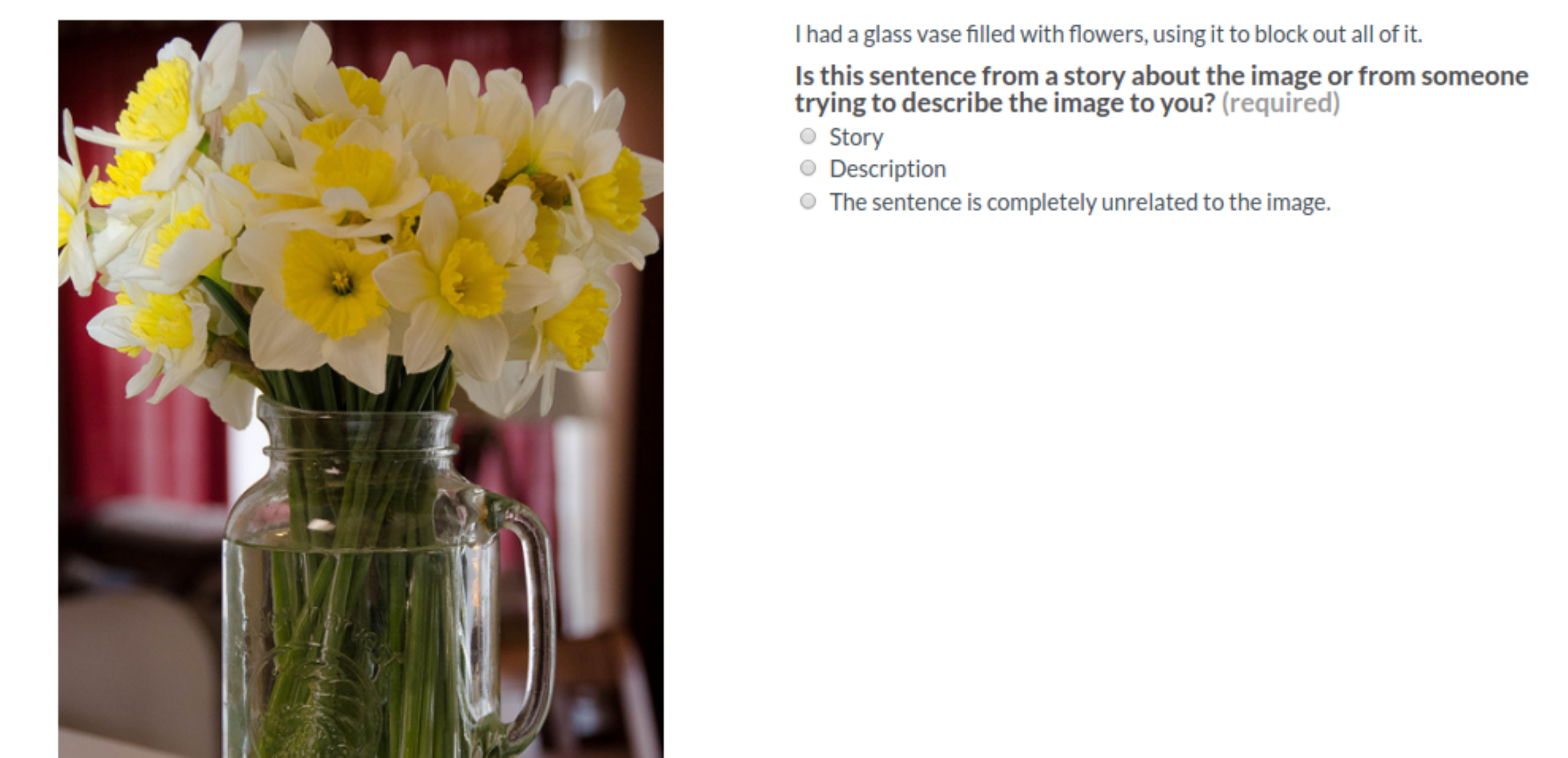}
\caption{A screen-shot of a single question asked of workers in the style evaluation task.}
\label{fig:heval_story_example}
\end{figure}

We performed two human evaluation tasks using the CrowdFlower\footnote{\url{https://www.crowdflower.com}} platform. The first was a relevance task, asking how well a caption describes an image on a four point scale. We provide screen-shots of the instructions given to workers, Figure~\ref{fig:heval_desc_inst}, and an example question, Figure~\ref{fig:heval_desc_example}. The second task evaluates conformity to the romantic novel style, by asking workers if the caption is from a story about the image, from someone trying to describe the image or completely unrelated to the image. We provide screen-shots of the instructions given to workers, Figure~\ref{fig:heval_story_inst}, and an example question, Figure~\ref{fig:heval_story_example}.

\subsubsection{Crowd-sourcing Quality Control and Rating Aggregation}

To ensure reliable results and avoid workers who choose randomly CrowdFlower injects questions with known ground truth into each task, requiring workers to achieve at least 70\% accuracy on these questions. We manually labelled a small selection of questions which were judged to be clear exemplars. On a limited number of our ground truth questions, workers consistently made mistakes. We revised or removed these question from the ground-truth. The ground truth was expanded by adding selecting questions to which all three annotators agreed on the answer. This is the method suggested by the CrowdFlower documentation for running large evaluations, because additional ground truth speeds up evaluation as workers may complete more tasks (ground truth is never re-used for the same worker and so acts as a limit on the number of tasks they can complete).

Each image-caption pair is seen by $n \ge 3$ workers. Where $n=3$ in most cases, typically being greater than 3 when workers have successfully challenged the original ground truth. We aggregate these judgements by assigning each one a weight $1/n$, and calculating the weight normalised sum for each possible answer. The resulting scores are displayed in Figure 3 of the main text. In the case of descriptiveness judgements a further summary statistic is calculated as the average descriptiveness score in the range 1-4.

\subsection{Results}
\label{sec:appendix_results}

\subsubsection{BLEU, METEOR, and CIDEr for styled captions}

Table~\ref{tab:semstyle_coco_cap_res} and Table~\ref{tab:semstyle_style_cap_res} provide additional automatic results, include BLEU, METEOR, and CIDEr scores -- as measured on the MSCOCO results. As we note in the main text these n-gram based measures are less relevant in the style generation case, but are provided here for completeness.

\begin{table}
\begin{center}
\footnotesize
\begin{tabular}{|l||c|c|c|c|c||c|c|c|}
\hline \textit{Model} & \textit{BLEU-1} & \textit{BLEU-4} & \textit{METEOR} & \textit{CIDEr} & \textit{SPICE} & \textit{CLF} & \textit{LM} & \textit{GRU LM}\\ \hline
\hline CNN+RNN-coco & 0.667 & 0.238 & 0.224 & 0.772 & 0.154 & 0.001 & 6.591 & 6.270\\ 
\hline StyleNet-coco & 0.643 & 0.212 & 0.205 & 0.664 & 0.135 & 0.0 & 6.349 & 5.977\\
\hline \mymodel-cocoonly & 0.651 & 0.235 & 0.218 & 0.764 & 0.159 & 0.002 & 6.876 & 6.507\\
\hline \mymodel-coco & 0.653 & 0.238 & 0.219 & 0.769 & 0.157 & 0.003 & 6.905 & 6.691\\ 
\hline
\end{tabular}
\caption{Evaluating caption descriptiveness
on MSCOCO dataset. For details of metrics see the main text for details of methods see Section~\ref{sec:baselines}.}
\label{tab:semstyle_coco_cap_res}
\end{center}
\end{table}

\begin{table}
	\footnotesize
	\centering
	\begin{tabular}{|c||c|c|c|c|c||c|c|c|}
	\hline \textit{Model} & \textit{BLEU-1} & \textit{BLEU-4} & \textit{METEOR} & \textit{CIDEr} & \textit{SPICE} & \textit{CLF} & \textit{LM} & \textit{GRU LM}\\ \hline
	\hline StyleNet & 0.272 & 0.099 & 0.064 & 0.009 & 0.010 & 0.415 & 7.487 & 6.830\\
	\hline TermRetrieval & 0.322 & 0.037 & 0.120 & 0.213 & 0.088 & 0.945 & 3.758 & 4.438\\   
	\hline neural-storyteller & 0.265 & 0.015 & 0.107 & 0.089 & 0.057 & 0.983 & 5.349 & 5.342\\ 
	\hline JointEmbedding & 0.237 & 0.013 & 0.086 & 0.082 & 0.046 & 0.99 & 3.978 & 3.790\\ \hline
	\hline \mymodel-unordered & 0.446 & 0.093 & 0.166 & 0.400 & 0.134 & 0.501 & 5.560 & 5.201\\ 
	\hline \mymodel-words & 0.531 & 0.137 & 0.191 & 0.553 & 0.146 & 0.407 & 5.208 & 5.096\\ 
	\hline \mymodel-lempos & 0.483 & 0.099 & 0.180 & 0.455 & 0.148 & 0.533 & 5.240 & 5.090\\ 
	\hline \mymodel-romonly & 0.389 & 0.057 & 0.156 & 0.297 & 0.138 & 0.770 & 4.853 & 4.699\\ 
	\hline \mymodel & 0.454 & 0.093 & 0.173 & 0.403 & 0.144 & 0.589 & 4.937 & 4.759\\ 
	\hline  
	\end{tabular}
	\caption{Evaluating styled captions with automated metrics. For \textit{SPICE} and \textit{CLF} larger is better, for \textit{LM} \& \textit{GRU LM} smaller is better. For metrics see the main text for baselines see Sec.~\ref{sec:baselines}.}
	\label{tab:semstyle_style_cap_res}
\end{table}

\subsubsection{Tabular Details for Human Evaluation}

Table~\ref{tab:human_eval_desc} and Table~\ref{tab:human_eval_story} give the full results for the human evaluation tasks. In the main text these are presented in graphical form, for completeness the full numerical results are given here.

\begin{table}[!ht]
\centering
\begin{tabular}{|c||c|c|c|c|}
\hline \textit{Method} & \textit{Desc 0} & \textit{Desc 1} & \textit{Desc 2} & \textit{Desc 3} \\ \hline
\hline CNN+RNN-coco & 15.6 & 16.7 & 24.2 & 43.4 \\
\hline neural-storyteller & 42.3 & 27.3 & 17.0 & 13.5 \\
\hline TermRetrieval & 24.4 & 28.5 & 20.3 & 26.8 \\ \hline
\hline \mymodel-romonly & 16.1 & 24.3 & 25.0 & 34.7 \\
\hline \mymodel & 12.2 & 23.2 & 20.9 & 43.8 \\
\hline 
\end{tabular} 
\caption{Human evaluations of the percentage of captions from each method that were, in regards to the image:
0 -- Completely unrelated, 1 -- Have a few of the right words, 2  -- Almost correct with a few mistakes, 3 -- Clear and accurate}
\label{tab:human_eval_desc}
\end{table}

\begin{table}[!ht]
\centering
\begin{tabular}{|c||c|c|c|}
\hline \textit{Method} & \textit{\% Unrelated} & \textit{\% Desc.} & \textit{\% Story}\\ \hline
\hline CNN+RNN-coco & 27.8 & 66.0 & 6.2 \\
\hline neural-storyteller & 44.2 & 3.2 & 52.6 \\
\hline TermRetrieval & 26.0 & 18.5 & 55.5 \\ \hline
\hline \mymodel-romonly & 21.6 & 24.5 & 53.8 \\
\hline \mymodel & 22.8 & 35.3 & 41.9 \\
\hline 
\end{tabular}
\caption{Human evaluations of the percentage of captions from each method that were judged as: unrelated to the image content, a basic description of the image, or part of a story relating to the image.}
\label{tab:human_eval_story}
\end{table}

\subsubsection{Hypothesis Tests for Human Evaluations}
\label{ssec:hypotest_human}

Statistical hypothesis testing (null hypothesis testing) for human story judgements is shown in Table~\ref{tab:semstyle_story_chitest}, for human descriptiveness judgements it is shown in Table~\ref{tab:semstyle_desc_chitest}. In both cases we have used $\mathcal{X}^2$ tests on method pairs with the beonferroni correction.

\begin{table}
\small
\centering
\begin{tabular}{|l||l|l|l|p{1.8cm}|}
\hline & CNN+RNN-coco & neural-storyteller & TermRetrieval & SemStyle-romonly \\ \hline
\hline \textit{CNN+RNN-coco} & - & - & - & - \\
\hline \textit{neural-storyteller} & 5.6e-09* & - & - & - \\
\hline \textit{TermRetrieval} & 1.2e-08* & 0.88 & - & - \\
\hline \textit{SemStyle-romonly} & 2.1e-12* & 0.18 & 0.13 & - \\
\hline \textit{SemStyle} & 1.4e-06* & 0.27 & 0.34 & 0.014 \\
\hline 
\end{tabular}
\caption{$\mathcal{X}^2$ tests on method pairs for \textbf{human story judgements}. We combine counts for ``unrelated" with ``purely descriptive", while ``story" is kept as its own class. Those marked with a * indicate rejection of the null hypothesis (H0: the two methods give the same multinomial distribution of scores) at p-value of 0.005 -- this is p-value of 0.05 with bonferroni correction of 10 to account for multiple tests.}
\label{tab:semstyle_story_chitest}
\end{table}

\begin{table}
\small
\centering
\begin{tabular}{|l||l|l|l|p{1.8cm}|}
\hline & CNN+RNN-coco & neural-storyteller & TermRetrieval & SemStyle-romonly\\ \hline
\hline \textit{CNN+RNN-coco} & - & - & - & - \\
\hline \textit{neural-storyteller} & 1e-56* & - & - & - \\
\hline \textit{TermRetrieval} & 4.1e-18* & 9.3e-14* & - & - \\
\hline \textit{SemStyle-romonly} & 0.00032* & 2.3e-35* & 3.4e-07* & - \\
\hline \textit{SemStyle} & 0.18 & 2.1e-48* & 1.7e-13* & 0.023 \\
\hline 
\end{tabular}
\caption{$\mathcal{X}^2$ tests on method pairs for \textbf{human descriptiveness judgements}. We combine counts for ``clear and accurate" with ``only a few mistakes", and ``some correct words" with ``unrelated". Those marked with a * indicate rejection of the null hypothesis (H0: the two methods give the same multinomial distribution of scores) at p-value of 0.005 -- this is p-value of 0.05 with bonferroni correction of 10 to account for multiple tests.}
\label{tab:semstyle_desc_chitest}
\end{table}

\subsubsection{Attributes of the Generated Style}
\label{ssec:attributes_genstyle}

The style of the text is difficult to define in its entirety, but we can look at a few easily identifiable style attributes to better understand the scope of the style introduced into the captions. First, we randomly sample 4000 captions or sentences from the MSCOCO and romance dataset. We then generate captions for 4000 images using \textit{CNN+RNN-coco} and \textit{CNN+RNN-coco}. On these four datasets we count: the percentage of sentences with past or present tense verbs (to identify the tense used in the captions), the fraction of sentences with first person pronouns (to identify sentences using first person perspective), the number of unique verbs used in the 4000 samples (to identify verb diversity). The results are summarised in Table~\ref{tab:semstyle_style_properties}. Parts-of-speech tags are obtained automatically with the spaCy\footnote{\url{https://github.com/explosion/spaCy/tree/v1.9.0}} library. For counting purposes, past tense verbs are those tagged with Penn Treebank tags VBD and VBN, while present tense verbs are those tagged with VBG, VBP and VBZ. Under this scheme gerunds and present participles are counted as present tense, while past participles are counted as past tense. Sentence may include verbs in both past and present tense, for example ``The dog was wearing a vest.", where ``was" is past tense and ``wearing" is present participle. Such sentences contribute to both the past and present tense counts.

Captions generated by \textit{\mymodel} use past-tense verbs in 75.0\% of sentences, which is close to the ground-truth level of 72.0\% and far greater than the descriptive method (\textit{CNN+RNN-coco}) at 10.6\%. This corresponds to a reduction in present tense verbs, consistent with the ground-truth. \textit{\mymodel} includes first person pronouns in 24.4\% of captions, compared to 0.0\% for \textit{CNN+RNN-coco}. The \textit{romance ground-truth} has personal pronouns in 31.2\% of sentences, which is higher than \textit{\mymodel} -- we expect that describing images limits the applicability of first person pronouns. \textit{\mymodel} has an effective verb vocabulary almost twice as large (92.3\% larger) as \textit{CNN+RNN-coco}, which suggests more interesting verb usage. However, both \textit{\mymodel} and \textit{CNN+RNN-coco} have lower verb diversity than either ground-truth dataset. We expect that some verbs that are not appropriate for image captioning and the RNN with argmax decoding tends to generate more common words. Compared to \textit{CNN+RNN-coco} the \textit{\mymodel} model reflects the ground-truth style by generating more captions in past tense, first person, and with greater verb diversity.
 
\begin{table}
\begin{center}
\small
\begin{tabular}{|p{2.8cm}||p{2.5cm}|p{2.0cm}|p{1.8cm}|p{1.8cm}|}
\hline  & Sentences with Present Tense Verbs & Sentences with Past Tense Verbs & Sentences with First Person Pronouns & Unique Verbs \\ \hline
\hline \textit{MSCOCO {ground-truth}} & 73.8\% & 17.0\% & 0.2\% & 497 \\ 
\hline \textit{romance ground-truth} & 51.4\% & 72.0\% & 31.2\% & 1286 \\ \hline
\hline \textit{CNN+RNN-coco}  & 70.4\% & 10.6\% & 0.0\% & 181 \\ 
\hline \textit{\mymodel} & 56.8\% & 75.0\% & 24.4\% & 348 \\ 
\hline 
\end{tabular} 
\caption{Statistics on attributes of style collected from 4000 random samples from two ground-truth datasets and 4000 test captions generated by the descriptive only model (\textit{CNN+RNN-coco}) and our \textit{\mymodel} model. We measure the fraction of sentences or captions with present tense verbs, past tense verbs or first person pronouns. We also count the number of unique verbs used in the sample.}
\label{tab:semstyle_style_properties}
\end{center}
\end{table}
 
To further explore the differences between styles we include Table~\ref{tab:semstyle_word_counts} that presents the most common lemmas for each dataset, stratified by POS tag. The most common nouns generated by \textit{\mymodel} have a greater overlap with the \textit{MSCOCO ground-truth} than the \textit{romance ground-truth}. This is the desired behaviour since nouns are a key component of image semantics and so nouns generated by the \termgen should be included in the output sentence. The most common verbs generated by \textit{\mymodel} are also similar to the \textit{MSCOCO ground-truth}; we expect this is a result of a similar set of common verbs in both ground-truth datasets. The use of determiners in \textit{\mymodel} more closely matches the \textit{romance ground-truth}, in particular the frequent use of the definite article ``the" rather than the indefinite ``a". The most common adjectives in all word sources typically relate to colour and size, and vary little across the different sources.

\begin{table}
\begin{center}
\small
\begin{tabular}{|r|l|}
\hline \multicolumn{1}{|c|}{Word Source} & Most Common Lemmas \\ \hline

\hline \multicolumn{1}{|l|}{\textit{MSCOCO ground-truth}} &  \\ 
\cline{1-1} NOUN & man(3.7\%), people(1.9\%), woman(1.8\%), street(1.5\%), table(1.4\%) \\ 
VERB & be(20.0\%), sit(9.3\%), stand(6.4\%), hold(4.4\%), ride(3.1\%) \\ 
ADJ & white(6.8\%), large(5.4\%), black(4.1\%), young(4.0\%), red(3.8\%) \\ 
DET & a(81.8\%), the(14.9\%), some(1.7\%), each(0.6\%), this(0.4\%) \\ \hline

\hline \multicolumn{1}{|l|}{\textit{romance ground-truth}} &  \\ 
\cline{1-1} NOUN & man(2.7\%), hand(1.5\%), eye(1.4\%), woman(1.3\%), room(1.2\%) \\ 
VERB & be(15.5\%), have(4.6\%), do(2.5\%), would(2.4\%), can(1.9\%) \\ 
ADJ & small(2.3\%), other(2.0\%), little(2.0\%), black(2.0\%), white(1.9\%) \\ 
DET & the(60.5\%), a(26.5\%), that(3.2\%), this(2.8\%), no(1.3)\% \\ \hline

\hline \multicolumn{1}{|l|}{\textit{CNN+RNN-coco}} &  \\ 
\cline{1-1} NOUN & man(6.9\%), group(3.0\%), people(2.6\%), table(2.6\%), field(2.3\%) \\ 
VERB & be(29.4\%), sit(15.4\%), stand(10.2\%), hold(5.6\%), ride(4.6\%) \\ 
ADJ & large(15.0\%), white(10.9\%), green(4.7\%), blue(4.5\%), next(4.5\%) \\ 
DET & a(91.9\%), the(7.7\%), each(0.2\%), some(0.1\%), an(0.1\%) \\ \hline

\hline \multicolumn{1}{|l|}{\textit{\mymodel}} &  \\ 
\cline{1-1} NOUN & man(5.5\%), table(2.8\%), street(2.7\%), woman(2.6\%), who(2.4\%) \\ 
VERB & be(24.5\%), sit(10.3\%), stand(4.8\%), have(3.6\%), hold(3.2\%) \\ 
ADJ & sure(14.7\%), little(9.4\%), hot(5.6\%), single(4.7\%), white(3.9\%) \\ 
DET & the(68.6\%), a(30.8\%), no(0.2\%), any(0.2\%), an(0.1\%) \\ \hline

\end{tabular} 
\caption{The most common words per part-of-speech category in the two ground truth datasets and in the sentences generated by the descriptive model (\textit{CNN+RNN-coco}) and \textit{\mymodel}. For each word we display the relative frequency of that word in the POS category -- represented as a percentage.}
\label{tab:semstyle_word_counts}
\end{center}
\end{table}

\subsubsection{Precision and Recall in the Semantic Term Space}
\label{ssec:pr_term_space}

\begin{table}
	\centering
	\begin{tabular}{|c||c|c|}
	\hline \textit{Model} & \textit{Precision} & \textit{Recall} \\ \hline
	\hline CNN+RNN-coco & 0.561 & 0.517 \\
	\hline StyleNet-coco & 0.506 & 0.468 \\ 
	\hline \mymodel-cocoonly & 0.636 & 0.531 \\
	\hline \mymodel-coco & 0.631 & 0.532\\ \hline
	\hline StyleNet & 0.027 & 0.028 \\
	\hline TermRetrieval & 0.505 & 0.336 \\   
	\hline neural-storyteller & 0.234 & 0.225 \\ 
	\hline JointEmbedding & 0.340 & 0.177 \\ \hline
	\hline \mymodel-unordered & 0.597 & 0.501 \\ 
	\hline \mymodel-words & 0.611 & 0.517 \\ 
	\hline \mymodel-lempos & 0.593 & 0.504 \\ 
	\hline \mymodel-romonly & 0.624 & 0.511\\ 
	\hline \mymodel & 0.626 & 0.517 \\ 
	\hline  
	\end{tabular}
	\caption{Precision (BLEU-1) and recall (ROUGE-1) in our semantic term space.}
	\label{tab:semstyle_pr_termspace}
\end{table}

To evaluate the precision and recall in the term space we match semantic terms in the output sentence with semantic terms in the caption ground truth. The results will depend on the efficacy of the visual concept detection pipeline (eg the \termgen for \textit{\mymodel}) as well as the language generation (eg the \langgen). While we expect a bias towards methods using our semantic term space, this analysis is useful for confirming \textit{\mymodel} accurately produces captions with term representations similar to the ground truth. Precision is reported as BLUE-1 without length penalty on terms, while recall is reported as ROUGE-1 on terms -- in both cases all ground truth reference sentences are used. BLEU-1 and ROUGE-1 are not effected by term ordering as they are uni-gram metrics. Results in Table~\ref{tab:semstyle_pr_termspace} shows that the four variants of \textit{\mymodel} (\textit{\mymodel-cocoonly}, \textit{\mymodel-coco}, \textit{\mymodel-romonly}, \textit{\mymodel}) that use our semantic term space perform better than other model variants and baselines not using term space. Demonstrating \textit{\mymodel} focuses on accurate reproduction of the semantic term space. The best performing models are \textit{\mymodel-cocoonly} with the largest BLEU-1 and \textit{\mymodel-coco} with the largest ROUGE-1 -- though both models score highly in BLEU-1 and ROUGE-1. This is in line with the other automatic metrics shown in Table~\ref{tab:semstyle_coco_cap_res}, although these metrics also show \textit{CNN+RNN-coco} is competitive. Of the baselines the best performing is \textit{TermRetrieval}, which retrieves romance sentences using query words from a \termgen (trained only on raw words in this case).

\subsubsection{Choosing Semantic Terms}
\label{ssec:explore_semantic_terms}

We defined the set of semantic terms by incorporating our domain knowledge, e.g. nouns are semantically important while determiners are not. Alternatively, we can learn which word classes carry semantic information.

We would like to know which word classes (adjectives, nouns, verbs , etc.) carry the most visually semantic information. Intuitively, we seek the word classes which, when removed, lead to the largest increase in entropy. One way to quantify this is the perplexity of the ground truth sentence after conditioning on input words belonging to different classes. For example, remove all nouns from the conditioning set of semantic terms and measure the change in perplexity. Balancing for class frequency is necessary, because removing unimportant words such as determiners could have a large effect on perplexity if they are frequent.

Our approach requires a probabilistic model with a domain including the word classes of interest and a range including possible output sentences. One, computationally expensive, solution is to train the language generation model for each possible word class. Instead we use a single language generation model trained on input sentences with 66\% of the input words randomly removed -- an approach reminiscent of de-noising auto-encoders. We train this model once and then selectively drop out words during testing.

Our search for the most important word classes, starts with uniform random removal of all words down to the 33\% level and thereby establishing a baseline. From there each possible word class is given a rank, higher ranked word classes are always completely removed before lower ranked word classes; removal stops when only 33\% of words remain. Words from classes of the same rank are chosen uniformly at random. For example if the input sentence is "the cat on the mat ." and the removal order had nouns ranked 2 and all other parts of speech ranked 1, then nouns "cat" and "mat" would both be removed. Remaining words would be randomly removed until only 2 out of the 6 remain. Using this method we should see the lowest perplexity when the words are ordered from least important to most important.

Our forward selection approach tries to set each word type to the highest non-occupied rank or the lowest non-occupied rank, the selection which minimises the perplexity is then fixed and the search proceeds until all classes are ranked. The final ordering was {\bf adjective, adverb, coordinating conjunction, particle, determiner, preposition or subordinate conjunction, verb, pronoun and noun}. With adjective judged the least useful and noun the most useful. Adjectives lack importance perhaps because they have only a local effect on a sentence and are often poorly detected by the CNN+RNN systems~\cite{Anderson2016,vinyals2015show}. This ordering is in line with our term space construction rules presented in the main paper.

\end{document}